\documentclass[letterpaper, 10 pt, conference]{ieeeconf}  

\IEEEoverridecommandlockouts                              
\usepackage{amsmath,amsfonts,amssymb,mathrsfs}
\usepackage{graphicx}
\usepackage{psfrag,graphicx,epsfig,epsf}
\usepackage[ruled,linesnumbered, noend]{algorithm2e}
\usepackage{fancyhdr}
\usepackage{wrapfig}
\usepackage{subfigure}
\usepackage{mathtools}
\usepackage[permil]{overpic}
\usepackage{url}
\usepackage{color}
\usepackage{verbatim}
\usepackage{xspace}
\usepackage{cite} 
\usepackage[export]{adjustbox}

\usepackage{multirow}

\usepackage{url}
\usepackage{breakurl}
\usepackage[breaklinks]{hyperref}

\makeatletter
\newcommand{\removelatexerror}{\let\@latex@error\@gobble}
\makeatother

\title{\LARGE \bf
Tight Robot Packing in the Real World\\ 
{\Large A Complete Manipulation Pipeline with Robust Primitives}
}

\author{Rahul Shome*$^+$, Wei N. Tang*, Changkyu Song,  Chaitanya Mitash, Hristiyan Kourtev,\\  Jingjin Yu, Abdeslam Boularias, and Kostas E. Bekris
\thanks{The authors are affiliated with the Computer Science Dept. of, Rutgers University, New Brunswick, and $^+$Rice University. Email:\{jingjin.yu,abdeslam.boularias,kostas.bekris\}@cs.rutgers.edu}
\thanks{*These authors have equally contributed to the work.}
\thanks{The authors would like to acknowledge the support of NSF 1723869, 1734492 and JD-X Research and Development Center (RDC). Any opinions expressed here or findings do not reflect those of the sponsor.}}

\begin{document}

\maketitle
\thispagestyle{empty}
\pagestyle{empty}
\newcommand{\danh}[2][1=]{\todo[linecolor=blue,
			backgroundcolor=blue!5,bordercolor=black,#1]{DH:#2}}
\newcommand{\kb}[2][1=]{\todo[linecolor=green,
			backgroundcolor=green!5,bordercolor=black,#1]{KB:#2}}
\newcommand{\ks}[2][1=]{\todo[linecolor=red,
			backgroundcolor=red!5,bordercolor=black,#1]{KS:#2}}
\newcommand{\rs}[2][1=]{\todo[linecolor=orange,
			backgroundcolor=orange!10,bordercolor=black,#1]{RS:#2}}
\newcommand{\jy}[2][1=]{\todo[linecolor=black,
			backgroundcolor=black!5,bordercolor=black,#1]{JJ:#2}}

\newcommand{\reals}{\mathbb{R}}
\newcommand{\integers}{\mathbb{Z}}

\newcommand{\Wspace}{\mathcal{W}}
\newcommand{\Objects}{\mathcal{O}}
\newcommand{\Manip}{\mathcal{M}}
\newcommand{\nobj}{k}

\newcommand{\Pspace}{\mathcal{P}}
\newcommand{\Pstable}{\mathcal{P}^s}
\newcommand{\pose}{p}
\newcommand{\GeomObj}{\mathcal{WO}}
\newcommand{\Arrange}{\mathcal{A}}
\newcommand{\Pumped}{\mathcal{A^P}}
\newcommand{\pumpedarr}{\alpha^{\mathcal{P}}}

\newcommand{\Qspace}{\mathcal{Q}}
\newcommand{\GeomManip}{\mathcal{WM}}

\newcommand{\Tspace}{\mathbb{T}} 
\newcommand{\Xspace}{\mathbb{X}}
\newcommand{\paths}{\Pi}

\newcommand{\roadmap}{\mathcal{R}}
\newcommand{\graph}{\mathcal{G}}
\newcommand{\nodes}{\mathcal{V}}
\newcommand{\node}{{v}}
\newcommand{\edges}{\mathcal{E}}
\newcommand{\edge}{{e}}
\newcommand{\prmstar}{{\tt PRM$^*$}}

\newcommand{\rpg}{${\tt RPG}$}

\newcommand{\local}{\mathcal{L}}

\newcommand{\prm}{{\tt PRM}}
\newcommand{\kprmstar}{{\tt k-PRM$^*$}}
\newcommand{\rrt}{{\tt RRT}}
\newcommand{\rrtdrain}{{\tt RRT-Drain}}
\newcommand{\rrg}{{\tt RRG}}
\newcommand{\est}{{\tt EST}}
\newcommand{\rrtstar}{{\tt RRT$^*$}}
\newcommand{\srrt}{{\tt RDG}}
\newcommand{\bvp}{{\tt BVP}}
\newcommand{\rdg}{{\tt RDG}}
\newcommand{\lrg}{{\tt LRG}}
\newcommand{\alg}{{\tt ALG}}
\newcommand{\upump}{{\tt UPUMP}}
\newcommand{\prxpump}{{\tt RPG}}
\newcommand{\fixed}{{\tt Fixed}-$\alpha$-\rdg}
\newcommand{\nrob}{k}
\newcommand{\cons}{K}

\newcommand{\frnodes}{V_f}
\newcommand{\frnode}{v_f}
\newcommand{\grnodes}{V_g}
\newcommand{\grnode}{v_g}
\newcommand{\fredges}{E_f}
\newcommand{\fredge}{e_f}
\newcommand{\gredges}{E_g}
\newcommand{\gredge}{e_g}
\newcommand{\kedges}{E_{\cons}}
\newcommand{\kedge}{e_{\cons}}
\newcommand{\safe}{q_s^{\mathcal{M}}}
\newcommand{\hedges}{E_H}
\newcommand{\hedge}{e_H}
\newcommand{\hnodes}{V_H}
\newcommand{\hnode}{v_H}
\newcommand{\hgraph}{H}
\newcommand{\nblank}{b}
\newcommand{\config}{C}
\newcommand{\cquery}{\mathbb{C}}
\newcommand{\pumped}{P}
\newcommand{\pumpedgraph}{\mathcal{G}_P}
\newcommand{\pnodes}{V_P}
\newcommand{\pnode}{v_P}
\newcommand{\pedges}{E_P}
\newcommand{\pedge}{e_P}
\newcommand{\signs}{\Sigma}
\newcommand{\sign}{\sigma}
\newcommand{\gsign}{\sigma_{\pumpedgraph}}
\newcommand{\cedges}{E_c}
\newcommand{\constraints}{\tt c}

\newenvironment{myitem}{\begin{list}{$\bullet$}
{\setlength{\itemsep}{-0pt}
\setlength{\topsep}{0pt}
\setlength{\labelwidth}{0pt}
\setlength{\leftmargin}{10pt}
\setlength{\parsep}{-0pt}
\setlength{\itemsep}{0pt}
\setlength{\partopsep}{0pt}}}
{\end{list}}

\newcommand{\dof}{{\tt DoF}}

\newcommand{\mam}{$\mathcal{G}_{\tt MAM}$}
\newcommand{\pr}{\ensuremath{\mathbb{P}}}

\newcommand{\rad}{\ensuremath{r(n)}}
\newcommand{\radstar}{\ensuremath{r^*(n)}}
\newcommand{\radi}{\ensuremath{r_i(n)}}
\newcommand{\radj}{\ensuremath{r_j(n)}}
\newcommand{\crossrad}{\ensuremath{r_R(n)}}
\newcommand{\crossradstar}{\ensuremath{r^*_R(n)}}
\newcommand{\impcrossrad}{\ensuremath{\hat r_R(n)}}
\newcommand{\allimpcrossrad}{\ensuremath{\hat r_{R}(n^R)}}
\newcommand{\ki}{\ensuremath{k_i(n)}}
\newcommand{\kj}{\ensuremath{k_j(n)}}

\newcommand{\mmgraph}{\ensuremath{\mathbb{G}}}
\newcommand{\mmgimp}{\hat\mmgraph}
\newcommand{\mmgexp}{\mmgraph}
\newcommand{\aograph}{\ensuremath{\mathbb{G}^{AO}}}
\newcommand{\tree}{\ensuremath{\mathbb{T} \ }}
\newcommand{\mmnodes}{\mathbb{\hat V}}
\newcommand{\mmedges}{\mathbb{\hat E}}
\newcommand{\mmnodestpprm}{\mathbb{V}_{\chi_i}}
\newcommand{\mmedgestpprm}{\mathbb{E}_{\chi_i}}
\newcommand{\mmnode}{\mathbb{\hat v}}
\newcommand{\mmedge}{\mathbb{\hat e}}
\newcommand{\sprmstar}{Soft-\ensuremath{ {\tt PRM} }}
\newcommand{\irs}{\ensuremath{ {\tt IRS} }}
\newcommand{\spars}{{\tt SPARS}}
\newcommand{\drrt}{\ensuremath{{\tt dRRT}}}
\newcommand{\drrtstar}{\ensuremath{{\tt dRRT^*}}}
\newcommand{\dadrrtstar}{\ensuremath{\tt da\_dRRT^*}}

\newcommand{\sig}{{\tt SIG}}
\newcommand{\rmaps}{\ensuremath{\mathfrak{R}}}

\newcommand{\mmprm}{\ensuremath{\text{Random-}{\tt MMP}}}
\newcommand{\astar}{{\ensuremath{\tt A^{\text *}}}}
\newcommand{\mstar}{{\tt M^{\text *}}}
\newcommand{\opens}{P_{Heap}}

\newcommand{\cost}{\textup{cost}}

\newcommand{\kiril}[1]{{\color{blue} \textbf{Kiril:} #1}}
\newcommand{\chups}[1]{{\color{red} \textbf{Chuples:} #1}}
\newcommand{\rahul}[1]{{\color{blue} #1}}
\newcommand{\changkyu}[1]{{\color{red} #1}}

\newcommand{\T}{\mathcal{T}}

\newcommand{\leftrm}{\ensuremath{\mathbb{R}_{l}}  }
\newcommand{\rightrm}{\ensuremath{\mathbb{R}_{r}}  }
\newcommand{\leftmetric}{\ensuremath{\mathbb{P}_{l}}  }
\newcommand{\rightmetric}{\ensuremath{\mathbb{P}_{r}}  }
\newcommand{\cfull}{\ensuremath{\mathbb{C}_{{\rm full}}}  }
\newcommand{\cfree}{\ensuremath{\mathbb{C}_{{\rm free}}}  }
\newcommand{\cobs}{\ensuremath{\mathbb{C}_{{\rm obs}}}  }
\newcommand{\cleft}{\ensuremath{\mathbb{C}_{{l}}}  }
\newcommand{\cright}{\ensuremath{\mathbb{C}_{{r}}}  }
\newcommand{\cshared}{\ensuremath{\mathbb{C}_{{s}}}  }
\newcommand{\cgoal}{\ensuremath{q_{{\rm goal}}}  }
\newcommand{\cstart}{\ensuremath{q_{{\rm start}}}  }

\newcommand{\gimpleft}{\ensuremath{\hat\mmgraph_l}}
\newcommand{\gimpright}{\ensuremath{\hat\mmgraph_r}}

\newcommand{\xrand}{\ensuremath{x^{\textup{rand} \ }}}
\newcommand{\xnear}{\ensuremath{x^{\textup{near} \ }}}
\newcommand{\xnew}{\ensuremath{x^{\textup{n}} \ }}
\newcommand{\xlast}{\ensuremath{x^{\textup{last} \ }}}
\newcommand{\xparent}{\ensuremath{x^{\textup{best} \ }}}

\newcommand{\lr}{\ensuremath{\mathbb{R}_{ls}}}
\newcommand{\rr}{\ensuremath{\mathbb{R}_{sr}}}
\newcommand{\lp}{\ensuremath{\mathbb{P}_{l}}}
\newcommand{\rp}{\ensuremath{\mathbb{P}_{r}}}

\newcommand{\motoman}{{\tt Motoman}}
\newcommand{\baxter}{{\tt Baxter}}
\newcommand{\ao}{{\tt AO}}

\newcommand\inlineeqno{\stepcounter{equation}\ (\theequation)}

\newcommand{\chomp}{\ensuremath{\tt CHOMP } }

\newtheorem{assumption}{Assumption}

\newcommand{\W}{\mathcal W}
\newcommand\perm[2][\^n]{\prescript{#1\mkern-2.5mu}{}P\_{#2}}
\newcommand\comb[2][\^n]{\prescript{#1\mkern-0.5mu}{}C\_{#2}}
\newcommand{\objectset}{\mathcal{O}}
\newcommand{\object}{o}
\newcommand{\workspace}{\mathcal{W}}
\newcommand{\taskspace}{\mathcal{T}}
\newcommand{\arrangement}{A}
\newcommand{\oar}{p}
\newcommand{\manipulators}{\mathcal{M}}
\newcommand{\manipulator}{\mathit{m}}
\newcommand{\arm}{m}
\newcommand{\taskset}{\mathcal{T}}
\newcommand{\task}{\mathit{T}}
\newcommand{\sol}{\Pi}
\newcommand{\state}{q}

\newcommand{\Aspace}{\mathcal{A}}
\newcommand{\Afree}{\mathcal{A}_{\rm val}}
\newcommand{\ainit}{A_{\rm init}}
\newcommand{\atarget}{\hat{A}_{\rm goal}}
\newcommand{\soma}{{\tt soma}}
\newcommand{\coma}{\ensuremath{{\omega}}}
\newcommand{\scoma}{\ensuremath{{{\Omega}}}}
\newcommand{\qset}{\mathcal{Q}}
\newcommand{\startq}{S}
\newcommand{\targetq}{T}

\newcommand{\act}{a}
\newcommand{\actset}{\mathbb{A}}
\newcommand{\trajset}{{\D}}
\newcommand{\moveset}{\bar{\mathcal{O}}}
\newcommand{\home}{Q}
\newcommand{\scomaset}{\{\scoma\}}
\newcommand{\tour}{{\Gamma}}
\newcommand{\tspgraph}{\graph_{\tour}}
\newcommand{\tspnodes}{\nodes_{\tour}}
\newcommand{\tspedges}{\edges_{\tour}}
\newcommand{\algo}{{\tt{TOM}}\xspace}
\newcommand{\kuka}{{\tt{Kuka }}}
\newcommand{\D}{D}
\newcommand{\sininv}{\sin^{-1}}
\newcommand{\cosinv}{\cos^{-1}}
\newcommand{\milp}{{\tt{MILP}}\xspace}
\newcounter{model}
\newenvironment{model}
{\refstepcounter{model}}
{\begin{center}
\textbf{Model. }~\themodel
\end{center}
\medskip}

\newcommand{\fk}{FK}
\newcommand{\ik}{IK}
\newcommand{\graspset}{\mathcal{G}}
\newcommand{\grasp}{\mathit{g}}
\newcommand{\vol}{\mathit{vol}}
\newcommand{\rigid}{\mathcal{R}}
\newcommand{\bin}{\mathcal{B}}
\newcommand{\binit}{\bin_{init}}
\newcommand{\btarget}{\bin_{goal}}

\newcommand{\amove}{\mathcal{A}} 
 \newcommand{\commentadd}[1]{{#1}}

\newcommand{\commentdel}[1]{{#1}}

\def\r#1{\textcolor{red}{#1}}
 \vspace*{-4mm}
\begin{abstract} Many order fulfillment applications in logistics, such
as packing, involve picking objects from unstructured piles before
tightly arranging them in bins or shipping containers. Desirable
robotic solutions in this space need to be low-cost, robust, easily
deployable and simple to control.  The current work proposes a
complete pipeline for solving packing tasks for cuboid objects, given
access only to RGB-D data and a single robot arm with a vacuum-based
end-effector, which is also used as a pushing or dragging finger. The
pipeline integrates perception for detecting the objects and planning
so as to properly pick and place objects.  The key challenges
correspond to sensing noise and failures in execution, which appear at
multiple steps of the process. To achieve robustness, three
uncertainty-reducing manipulation primitives are proposed, which take
advantage of the end-effector's and the workspace's compliance, to
successfully and tightly pack multiple cuboid objects. The overall
solution is demonstrated to be robust to execution and perception
errors. The impact of each manipulation primitive is evaluated in
extensive real-world experiments by considering different versions of
the pipeline. Furthermore, an open-source simulation framework is
provided for modeling such packing operations. Ablation studies are
performed within this simulation environment to evaluate features of the proposed primitives.
\end{abstract}
 
\section{Introduction}
\label{section:introduction}
\begin{figure}[t]
\centering
\includegraphics[height=0.9in]{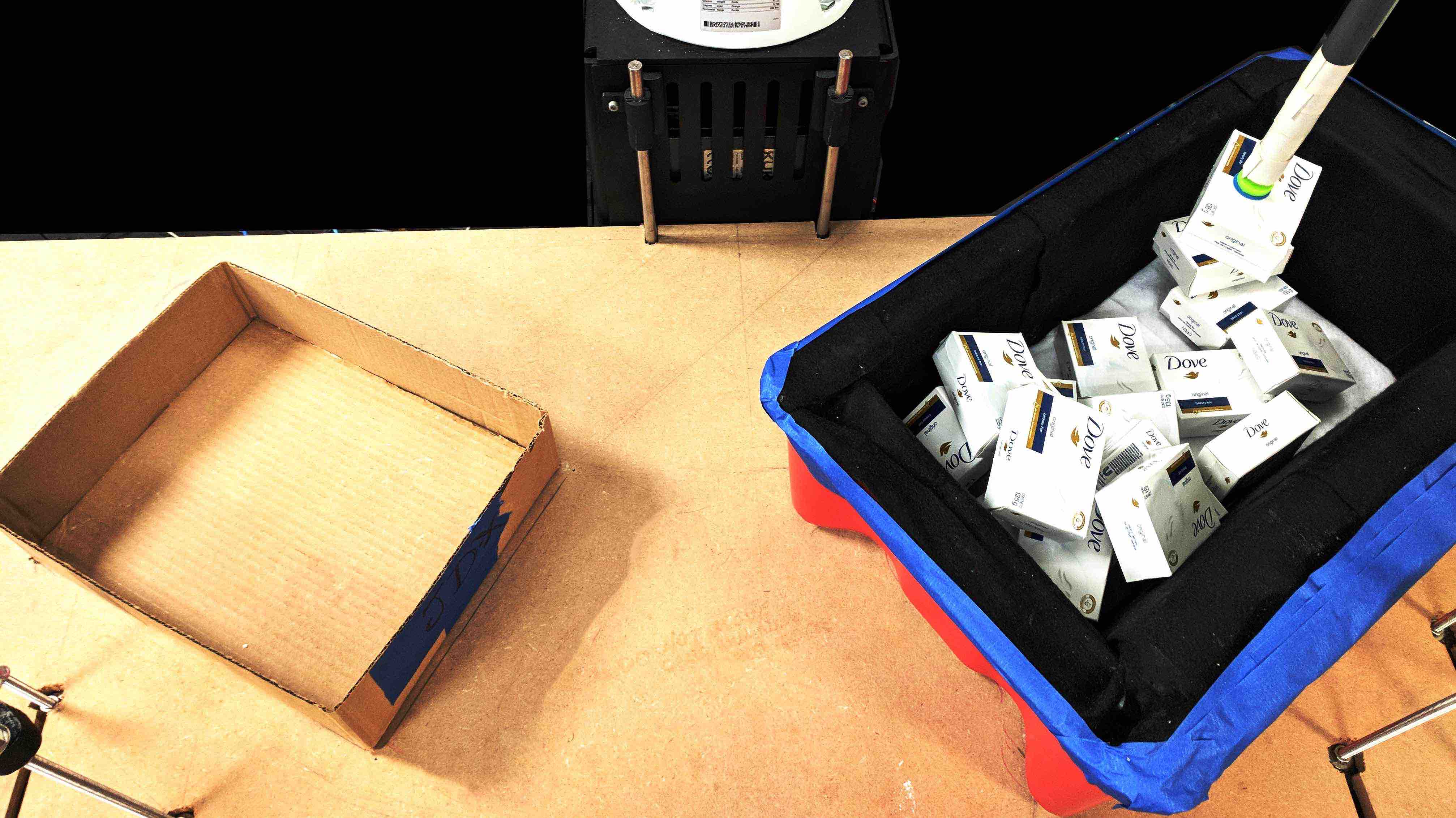}
\includegraphics[height=0.9in]{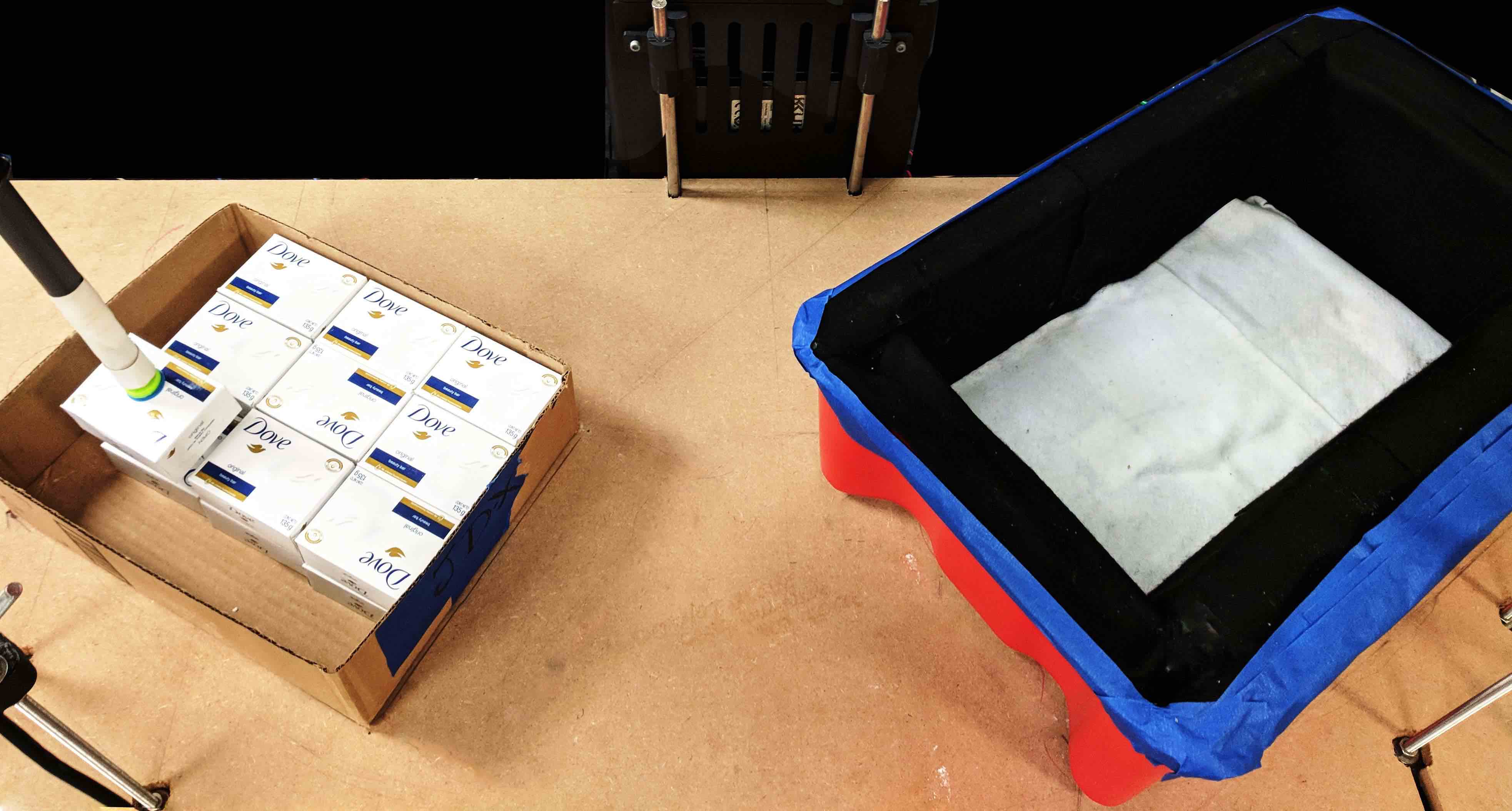}\\
\vspace{0.05in}
\includegraphics[height=1.75in]{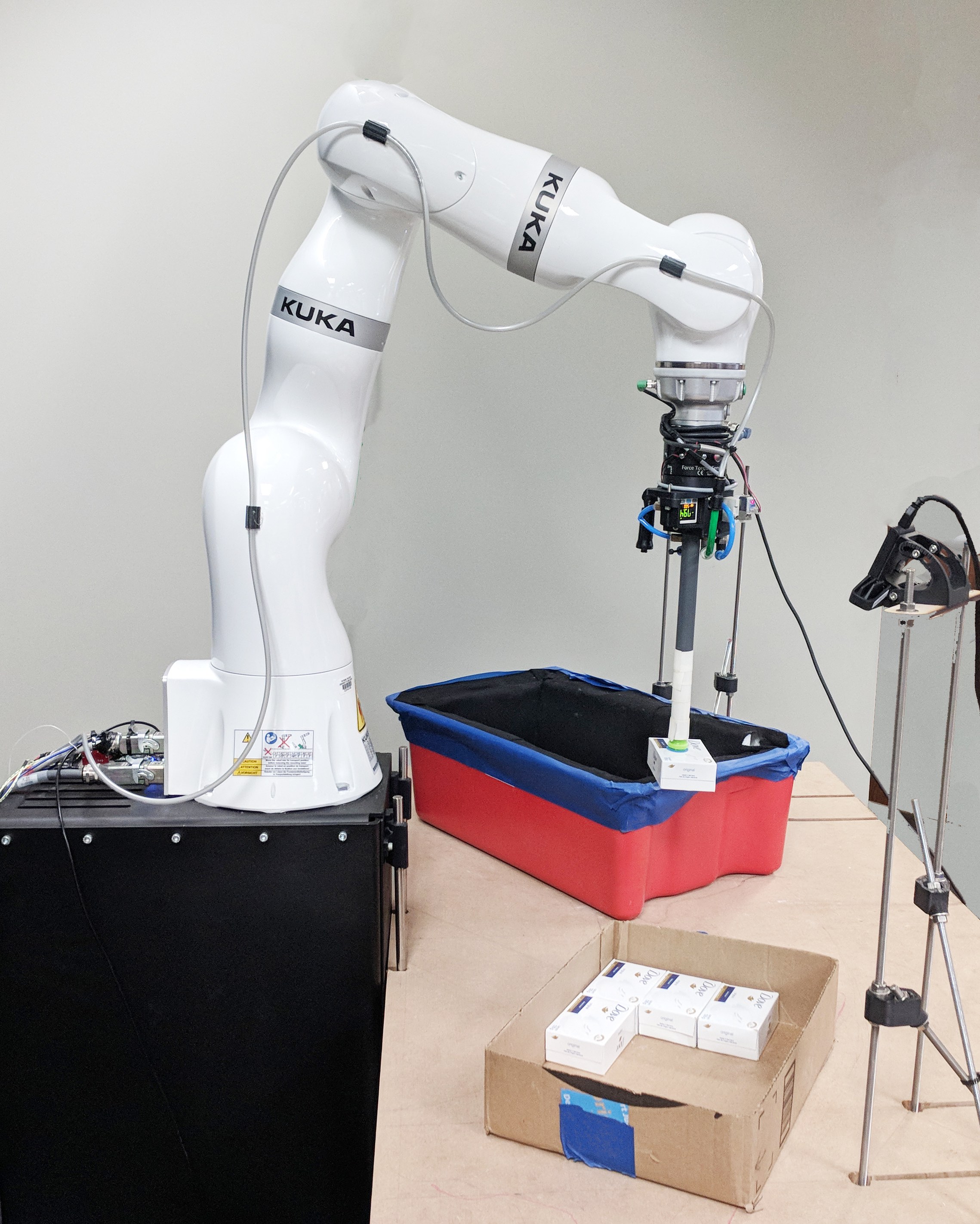}
\includegraphics[height=1.75in]{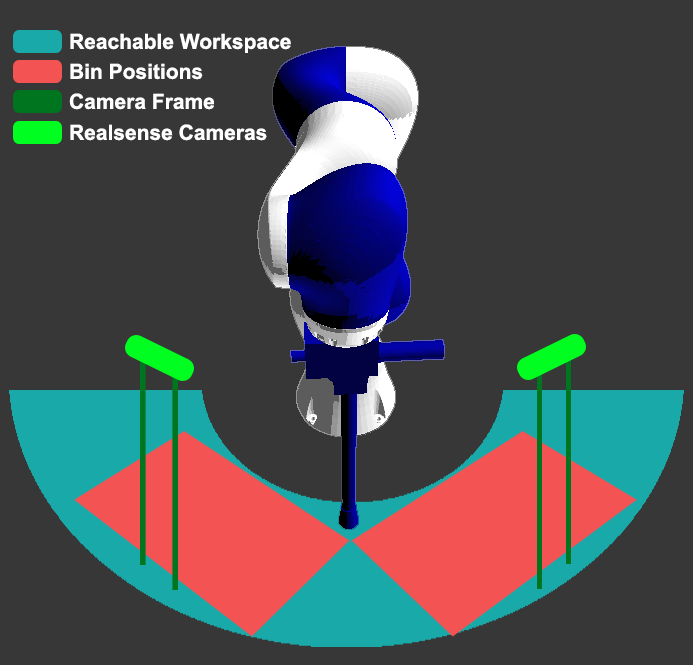}
\vspace{-0.1in}
\caption{(\textit{Top row:}) Tight packing for cuboid products: initial
configuration (left), and achieved packing (right). (\textit{Bottom
left:}) The problem is solved using a {\it Kuka LBR iiwa} arm equipped
with a suction-based end-effector and depth-sensing cameras {\it
SR300}. (\textit{Bottom right:}) Two bins are placed within the
reachability region of the arm given overhand picks.}
\label{fig:new_setup}
\vspace*{-0.25in}
\end{figure}

The past decade has witnessed a growth of autonomous robot
solutions for logistics and warehouse automation, such as the {\it
Amazon/Kiva} fulfillment system
~\cite{Enright:2011:OCA:2908675.2908681}.  Many tasks, however,
still rely on repetitive human labor, such as building customer
orders. In particular, picking packaged products from unstructured
piles and tightly packing them into another container is a frequent
task but often is performed
manually~\footnote{The authors would like to thank Dr. Hui Cheng and
members of the JD-X Research and Development Center for introducing
the packing task.}. This is because packing objects in confined
spaces, such as shipping boxes, as in Fig.~\ref{fig:new_setup} (top),
is challenging. It requires placing objects in close vicinity to each
other, in an ordered manner and aligned to the container's
boundary. This demands high accuracy both from perception and
manipulation. Apart from recent insights into the combinatorial, and
geometric bottlenecks~\cite{wang2019stable,wang2019robot}, relatively
little work exists that addresses real-world complexities of this
task, let alone using commodity hardware, such as a robotic arm, a
suction-based end-effector and an RGB-D sensor.

\noindent {\bf Warehouse Automation:} This article is
motivated by real-world warehouse automation tasks. Most
automation efforts so far have focused on picking. Product
placement is also critical but less automated\footnote{See an online
survey of industrial warehouse
automation here:\\ \url{https://robotpacking.org/industrial_automation.html}}.
Examples of placement tasks include: (i) \textit{Induction:} This relatively easier task requires dropping a picked object in an unobstructed
surface (e.g., conveyor), potentially with a desired
orientation. (ii) \textit{Specialized Packaging:} This often involves
mechanisms that build the container around the product. The objects often need to be properly pre-positioned. (iii) \textit{Stowing:} Placing a product in a
container for storage without regard to its final pose. (iv) \textit{Aggregation:} Bringing together batches of identical products (depanning) or containers (denesting). Palletizing requires containers to be packed into pallets with some degree of uniformity and careful positioning. (v) \textit{Tight Packing:} The
focus task of this work requires precise positioning of objects and
containers. It can be addressed without sensing if the object and
container positions are guaranteed by a conveying mechanism, which is
not always available and requires additional investments. Robust, and
efficient autonomous packing can provide significant economic, and
environmental savings by reducing packaging material, storage space,
and shipping costs. These costs typically form a good fraction of expenses in order-fulfilment and online retail.

\noindent {\bf Contribution:} This work develops a robot packing pipeline using
a minimalistic end-effector for different cuboid products presented in an unstructured pile that must be placed in arbitrarily sized open boxes. The focus is on making this process robust for real-world deployment. To help narrow the application gap and enable the reliable, fully autonomous execution of packing, this article:

\noindent \emph{A. Proposes minimalistic hardware and accompanying software architecture}. The hardware shown in Fig.\ref{fig:new_setup} (bottom-left) integrates a single robot arm, depth-imaging technology and a suction-based end-effector, which is also used as a pushing or dragging finger. While objects can be initially presented in configurations that require reorientation for proper placement, the suction-based end-effector is shown sufficient to solve such challenges. The result is a fully autonomous integrated system, as shown in Fig.~\ref{fig:new_setup} (top-right).

\noindent \emph{B. Develops corrective manipulation primitives to increase robustness} due to uncertainty arising both from actuation errors and noise in visual sensing. These closed-loop primitives intentionally use \textit{contacts} and \textit{compliance} of objects,  bins, and the end-effector to resolve the following tasks in real-time: (i) object \emph{toppling} in the unstructured bin to expose a desirable object surface for picking; (ii) \emph{adaptive pushing} of objects towards their target placement while displacing neighboring objects to further pack them given point cloud data; and (iii) real-time monitoring of potential failures and \emph{fine correction} to achieve tight packing.

\noindent \emph{C. Provides an open-source simulation framework for robot packing}. The simulation framework is made available to the research community to facilitate benchmarking in this space\footnote{\url{https://robotpacking.org/simulator.html}}.  The simulator
allows the modeling of high-fidelity RGB-D sensor information and compliant interactions.  

\noindent \emph{D. Extensively evaluates the pipeline and its primitives} both in simulation and with the real platform of Fig. \ref{fig:new_setup} (bottom-left). The experiments show that the proposed primitives provide robustness despite the setup's minimalism.

\section{Related Work}
\label{section:related}
\begin{figure*}
\centering
\includegraphics[width=1\textwidth]{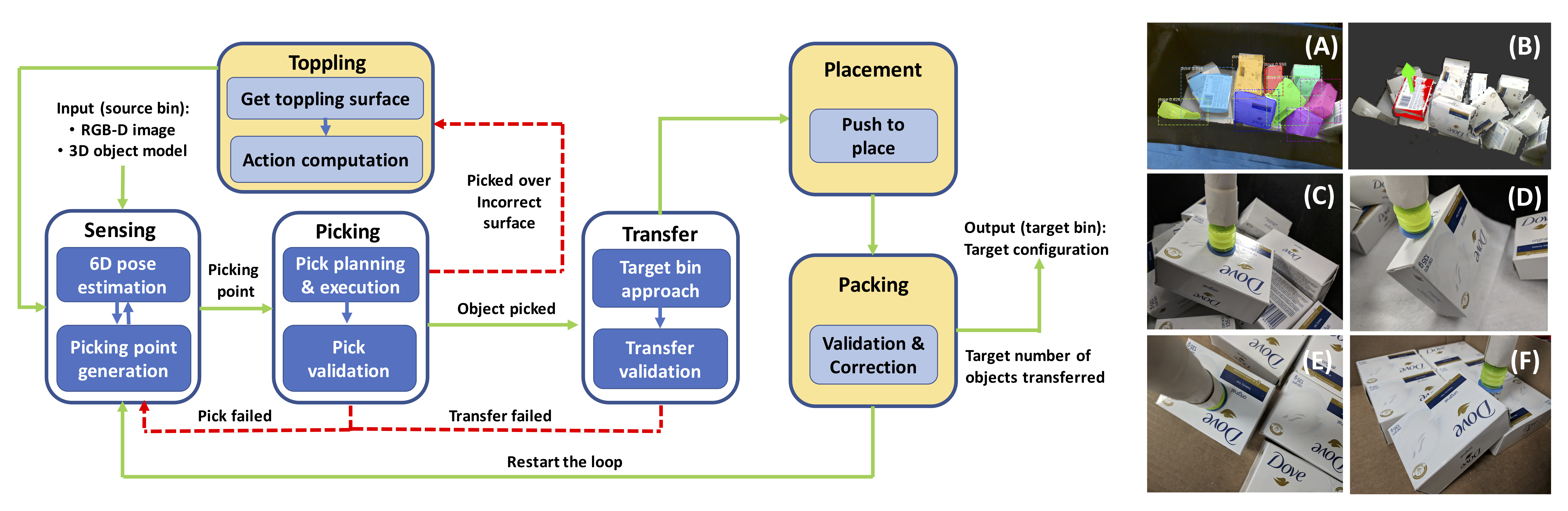}
\vspace{-0.35in}
\caption{\textit{Left}: Pipeline with the system modules (blocks)
highlighting data flow (green lines) and failure handling (red lines).
Sensing receives an RGBD image and object CAD models to return a grasp
point. Depending on the picking surface, the object is either
transferred to the place scene or is handled by the {\tt Toppling}
module, which flips the object in the pick scene. When the object is
transferred, a robust {\tt Placement} module places the object at the
target pose. The {\tt Packing} module validates and corrects the placement
to achieve tight packing. \textit{Right}: a) Instance segmentation. b)
Pose estimation and pick point selection, c) Picking d) Toppling e)
Placement and f) Packing.}
\vspace{-0.2in}
\label{fig:pipeline}
\end{figure*}

This work lies at the intersection of robot manipulation, perception,
and solutions for industrial deployment.

\noindent \textbf{Picking Objects in Clutter:} Traditionally, grasping used fingered hands~\cite{Sahbani:2012:OOG:2109688.2109859} for form- or force-closure for standalone objects~\cite{doi:10.1177/027836499601500302} or for clutter~\cite{Bohg:2014:DGS:2714095.2714355,
Boularias:2015:LMU:2887007.2887192, kimmel2020}. Suction-based grippers were shown successful in the Amazon Picking
Challenge \cite{correll2016analysis, schwarz2018fast,
morrison2018cartman, zeng2018robotic, 166, rennie_dataset} and attracted more interest. Model-based or data-driven grasping~\cite{Bohg:2014:DGS:2714095.2714355} often identifies points on object surfaces and hand poses that result in grasps. Model-based methods rely on object models to identify picks that are stable. They can be sensitive to modeling errors. Data-driven techniques rely on examples of grasps to regress successful
grasps ~\cite{DBLP:journals/corr/abs-1804-05172} without depending
on models. They tend to be computationally efficient during
inference~\cite{mahler2017binpicking} but require a sufficient amount of training data. This work follows a model-based approach using suction and aims for simplicity, robustness and reduced computational cost.

\noindent \textbf{Non-prehensile Manipulation:} Actions beyond grasping,
such as pushing, are helpful in manipulating objects in
clutter, such as for reducing the uncertainty of a target object's
pose~\cite{Dogar-2011-7322,Dogar22012} and for
rearrangement~\cite{king2015nonprehensile, cosgun2011push}. The
current work uses non-prehensile manipulation to topple objects,
inspired by early work \cite{lynch1999toppling}, and for placement to
counter the effects of inaccurate object localization.  A related
approach~\cite{chavan2015prehensile} performs within-hand manipulation
of an object by pushing it against its environment. The proposed
system takes advantage of the end-effector's compliance and leverages
contact with the environment to accurately place objects or topple
them. Recent work is exploring robust toppling~\cite{correa2019robust}, and the modeling of compliant contacts~\cite{cheng2019manipulation}.

\noindent \textbf{6D Pose Estimation:} A lot of
recent efforts in 6D pose estimation employ deep learning by performing regression over object orientations and centers from
images~\cite{xiang2017posecnn}. An alternative approach first predicts 3D
object coordinates, followed by a RANSAC scheme to predict the
object's pose~\cite{brachmann2014learning}. Geometric consistency has
also been used to refine the predictions from learned
models~\cite{michel2017global}. The current work is based on the authors' prior effort~\cite{stocs}, which first performs image segmentation to then guide a geometric search process for estimating 6D poses of objects~\cite{narayanan2016discriminatively}. Uncertainty over pixel labels is returned by a convolutional neural network and used to register 3D
models into the point cloud. Recent progress, motivated by the packing application, deals specifically with pose estimation of regular arrangements of similar objects closely situated in a container~\cite{mitash2019scene}.

\noindent \textbf{Bin Packing Algorithms:} This NP-hard
problem~\cite{Berkey1987, Martello:2000} considers the spatial
arrangement of objects of different or similar volumes so that they
are packed into a cubical bin. Most strategies search for
$\varepsilon$-optimal solutions via greedy
algorithms~\cite{Albers:1998:AAF:314613.314718}.  To the best of the
authors' knowledge, these solutions are not deployed in real
robotic workcells, where inaccuracies in vision and control
must also be dealt with.  Recent work~\cite{wang2019stable,
wang2019robot} has focused on computing incremental stable
arrangements for packing  objects in a
container. Such methods could be integrated with the proposed pipeline
to specify target arrangements.

\noindent {\bf Relation to Previous Work by the Authors:} This article extends previous work~\cite{shome2019towards} as follows: \emph{A.} The simulator is a new contribution. New ablation experiments using the simulator test: 1) the robustness of the primitives against different levels of simulated noise, and 2) the
impact of underlying parameters. \emph{B.} A novel extension to the adaptive pushing primitive solves problems where there is no clearance for the target objects given sufficient compliance. \emph{C.} A real world demonstration is included in Section~\ref{section:evaluation} where the containers' size is exactly the size of the packed arrangement. \emph{D.} Additional demonstrations performed on the robot show new capabilities of the proposed system: 1) toppling the object to its narrow, less-stable side, and 2) handling a pile with different objects and packing them into separate boxes. \emph{E.}  The text expands upon the primitives' technical details in Section~\ref{section:solution}, including pseudocode,  and the coverage of the related literature in Section~\ref{section:related}.

\section{Problem Setup and Notation}
\label{section:problem}
Consider a workspace $\Wspace$ with: a) a robot arm, b) static
obstacles, c) $n$ movable cuboid objects $\objectset=\{ \object^1, \ldots, \object^n \}$, d) two static container bins $\binit$ and $\btarget$, which are compliant bodies in known poses that define cuboid volumes, where the objects can be placed.

A labeled arrangement $\arrangement = \{ \pose^1, \ldots, \pose^n \}$ is an assignment of poses $\pose^i \in SE(3)$ to each object $\object^i$. The objects start at an initial, random but ``stable'' arrangement $\ainit$ inside $\binit$, i.e., the objects are stably resting and not moving. The $\ainit$ arrangement is \emph{not} known a-priori to the robot. The objective is to move $\objectset$ to an unlabeled arrangement $\atarget = \{ \hat{\pose}^1, \ldots, \hat{\pose}^n \}$, which achieves a tight packing in $\btarget$. $\atarget$ depends on the pose of $\btarget$, its dimensions and the objects' dimensions. The target unlabeled arrangement is input for the proposed process. $\atarget$ is a regular grid packing for the cuboid objects, which maximizes the number of objects inside  $\btarget$, as they rest on a stable face, and minimizes the convex hull of their volume. The unlabeled nature of $\atarget$ means it is satisfied as long as one of the objects is placed at each target pose, i.e.,

\vspace*{-.25in}
\begin{equation}
\forall \hat{\pose}^j \in \atarget: \exists\  \object^i \in \objectset \textrm{ so that } D(\hat{\pose}^j, p^i) < \epsilon,
\vspace*{-.1in}
\label{eq:satisfaction}
\end{equation}

\noindent where $\epsilon$ is a threshold for achieving the target pose; $D(\cdot,\cdot)$ is distance between object poses, which considers the 3-axis symmetry of the cuboid objects, i.e., if two poses result into an object occupying the same volume, their distance is 0. Such a distance metric for 6D poses is the ADI metric ~\cite{hinterstoisser2012model}.

The arm has $d$ joints that define the arm's configuration space $\cfull \subset \mathbb{R}^d$, which has a subset $\cfree$ that is collision-free given the static obstacles and the bins. Valid arm motions correspond to a continuous C-space curve $\pi : [0,1] \rightarrow \cfree$.  The arm has an end-effector, such as a suction cup, for which discrete operations $\{ \mathtt{pick}, \mathtt{release} \}$ give rise to discrete manipulation modes: $\mathcal{M} = \{ \mathtt{transfer}, \mathtt{transit} \}$. No within hand manipulation operations are available. The state space of the arm is: $\mathcal{X} = \cfree \times \mathcal{M}$. Sensing is used to reason about the current object poses. Overall, the \emph{robot operations} involve (i) rotating the joints, (ii) picking or releasing objects and (iii) sensing.

The arm's forward kinematics define a mapping $\fk: \cfull \rightarrow SE(3)$, which provides the pose $g \in SE(3)$ of the end-effector given $q \in \cfull$. The reachable task space defines the set of end-effector poses that can be reached without collisions: $\mathcal{T} = \{ \forall\ q \in \cfree: FK(q) \in SE(3)\}.$ For the arm to pick $\object^i$ at $\pose^i$, it has to be that the end-effector's pose $\grasp$ satisfies a binary output function: $\mathtt{is\_pick\_feasible}( \object^i, \pose^i, \grasp), \textrm{ where } \grasp \in \mathcal{T}.$ For instance, the pose $\grasp$ of a suction cup must align with one of the surfaces of an object $\object^i$ at $\pose^i$. Then, the set of end-effector poses that allow to pick an object at a specific pose are: $$\graspset( \object^i, \pose^i ) = \{\grasp \in \mathcal{T}:  {\mathtt{is\_pick\_feasible}}(\object^i, \pose^i, \grasp) = true\}.$$  Assume $\graspset( \object_i, \pose_i )$ is non-empty for all objects $\objectset$ and poses in $\ainit$ or $\atarget$. Otherwise, the task is not feasible. Note it may be necessary to reconfigure the objects inside $\binit$ to pick them from an appropriate face before placing them, given the lack of within-hand manipulation. Then, the task is to identify a sequence of \emph{robot operations} to transfer objects  $\objectset$ from the unknown initial stable arrangement $\ainit$ in $\binit$ to a tight, grid-based packing inside $\btarget$ that satisfies Eq. \ref{eq:satisfaction} given an unlabeled arrangement $\atarget$.

\section{Proposed Solution}
\label{section:solution}
\rahul{
\begin{figure*}[h]
    \centering
    \includegraphics[width=0.8\textwidth]{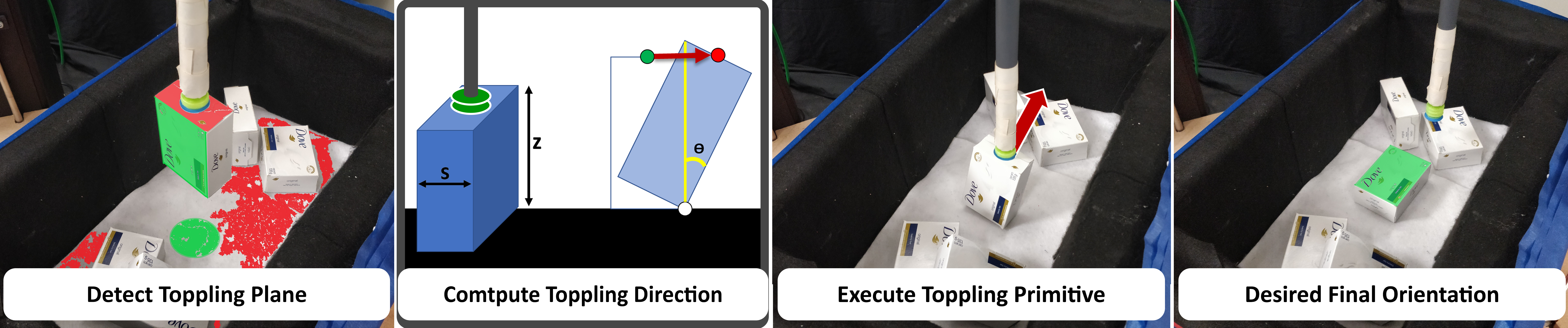}
    \vspace{-0.1in}
    \caption{Steps for toppling when the object's pose does not directly allow the desired placement configuration: (from left to right) a plane is detected in the scene that can allow toppling; the toppling direction is computed, where \textit{z} is the height of the face to be picked; toppling is executed by moving the end-effector along a vector of motion for the top face; the reoriented object is picked from a pose that allows the desired placement.}
    \vspace{-0.2in}
    \label{fig:toppling_steps}
\end{figure*}
}

Fig.~\ref{fig:pipeline} describes a pipeline for solving packing given the setup of Fig.~\ref{fig:new_setup}. More details about the hardware are provided in Section \ref{section:evaluation}. A straightforward baseline involves: a) {\em "Sensing"} objects  $\objectset$ in $\binit$ and selecting an object $o^i$; b) {\em "Picking"} $o^i$ by executing action $\{\tt pick\}$; c) "Transfer" of $o^i$ to the next available target pose $\hat{p}^j$ in $\btarget$ by executing $\{\tt release\}$ at that pose.  Experiments show that this baseline performs poorly given pose uncertainty, calibration errors, object non-uniformity and unexpected contacts. This motivates remedies, which actively increase robustness. To this end, three manipulation primitives are designed to increase robustness and are integrated with the overall architecture: i) {\em "Toppling"}; ii) {\em "Adaptive Pushing"}; and iii) {\em "Corrective Actions"}. This section describes first a baseline and then each one the proposed robust manipulation primitives.

\subsection{Baseline: Sense, Pick and Transfer}

Given an RGB-D image of the source bin $\binit$ and a CAD model of the target object category, the objective is to select $\object^i$ that can achieve a pose $\pose^i$ given the next available target pose $\hat{p}^j$ so that $D(p^i,\hat{p}^j) \leq \epsilon$. To achieve this, the image is passed through {\tt Mask-RCNN}~\cite{he2017mask} trained to perform segmentation and retrieve the visible subset of object instances $\objectset$.  An image segment is ignored if it has a number of pixels below a threshold or if {\tt Mask-RCNN} has small confidence the segment is of the target object category. 

The remaining segments are arranged in a descending order given the mean global Z-coordinate of all the RGB-D pixels in each segment. Then, 6D pose estimation is performed for the highest-order instance~\cite{175}\cite{stocs}.  If the detected 6D pose reveals that the object's top-facing surface does not allow its placement via a top-down pick and place, the next segment instance in the ordering is evaluated. If no object reveals a top-facing surface, then the object with the maximum mean global Z-coordinate is chosen for picking. 

For the chosen object, a picking point, i.e., a point where the suction cup is attached, is computed over the point cloud registered against the object model. The approach uses a pre-computed picking score on the object model, which indicates the pick's stability by minimizing the distance to the object's center. A continuous neighborhood of planar pickable points are required to make proper contact between the suction cup and the object's surface. A local search is performed around the best score point to maximize its pickable surface. The pick is performed at that maximum and the picked object is transferred to the target pose.

\subsection{Toppling}
The toppling primitive is invoked if there exists no object, which exposes the desirable top-facing surface, or if the object was erroneously picked from the wrong face. The latter is detected after the pick by performing pose estimation once the object is attached to the suction cup. For instance, this can happen for the soaps in Figure \ref{fig:pipeline} (right)(D), if only the thin side is available for pick but the soap needs to be placed on its wide side. In these cases, toppling is performed to reorient the object. Given the starting pose of an object $\pose_{start}$ and a toppling action of the arm, the object ends up at pose $\pose_{topple}$.  The requirement is for  $\pose_{topple}$ and the next available target pose $\hat{p}^j$ to have the same top-facing surface.

Prior work \cite{lynch1999toppling} has shown the efficacy of minimal end-effectors used in tandem with the environment to achieve toppling. In the previous work, the friction against a conveyor belt is used to topple an object about a resting surface. The conveyor belt's motion is parallel to the initial resting surface plane. In the current setup, the compliance of the suction cup is used to emulate the same effect using a lateral motion on the same plane as the top-surface along the direction of the desired transformation between $\pose_{start}$ and $\pose_{topple}$. Due to the symmetry of cuboidal objects, at least one neighboring surface allows a successful toppling to exist. The accompanying results show this to be highly effective in the target setup.

Algorithm~\ref{algo:topple} outlines the toppling process. It returns a sequence of poses, which constitute the discrete steps of the toppling maneuver. When executed, they drop the object to expose a desired face, denoted by the target orientation. It is invoked once the object is already attached to the end-effector using a top-down pick, and is also visible to the camera. Line 2 checks if reorientation is necessary by measuring the distance between the current object rotation $r$ and the target rotation $r_{\mathrm target}$. The toppling plane is computer over the point cloud based on the dimensions of the object's bottom surface. $\pose_{\mathrm plane}$ denotes the new object pose when its bottom surface touches the toppling plane.

\vspace{-.15in}
 \begin{algorithm}
 \small
 \DontPrintSemicolon
 \KwIn{Point cloud $X$, object $\object$, top-down current object pose $\pose_{start}\gets(t,r)$, target orientation $r_{\mathrm target}$}
 \KwOut{Toppling primitive $\Pi$  }
 	$\Pi\gets\emptyset$;\\
 	\If{ $D$($\pose_{start}, (t,r_{\mathrm target}$)) $\ >\ \epsilon $ }
 	{
 		$t_{\mathrm plane} \gets ${\sc GetTopplingPlane}($X, ${\sc Dimensions}($\object$));\\
 		$\pose_{\mathrm plane} \gets (t_{\mathrm plane}, r)$\\
 		$\Pi \gets \Pi \cup \pose_{\mathrm plane}$;\\
 		$(\theta,\Delta x,\Delta z) \gets ${\sc GetOffset}($\object, \pose_{\mathrm plane}, r_{\mathrm target}$);\\
 		$\Pi \gets \Pi \cup $ {\sc ApplyOffset}($\pose_{\mathrm plane}, \theta,\Delta x,\Delta z$);
 	}
 \Return{$\Pi$}
 \caption{{\sc Topple}}
 \label{algo:topple}
 \end{algorithm}
 \vspace{-.15in}
 
 \begin{figure*}[t]
\centering
\begin{overpic}[height=1.2in]{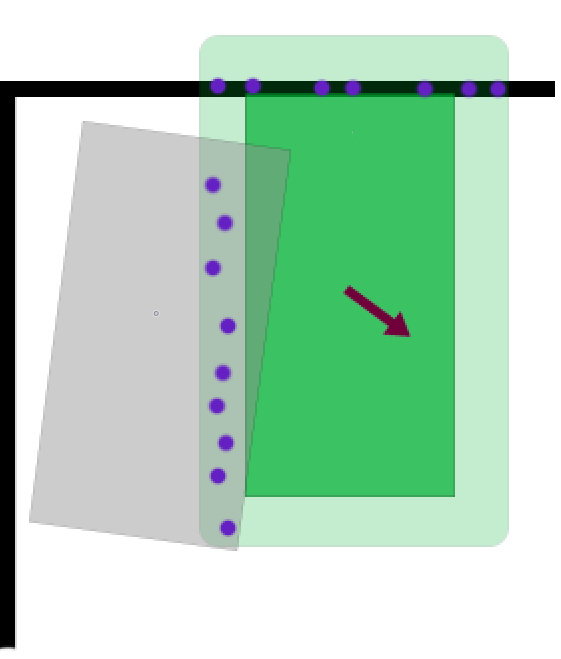}
\put(350,-100){(a)}
\end{overpic}
\begin{overpic}[height=1.2in]{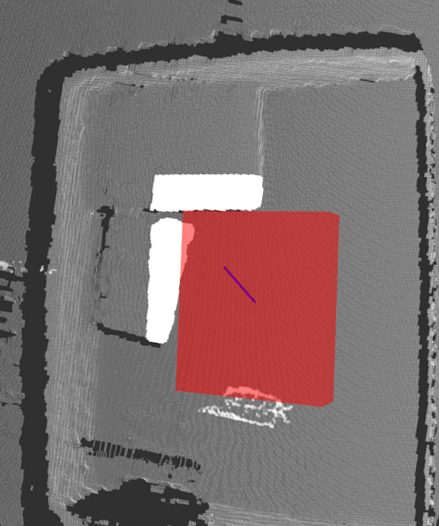}
\put(350,-100){(b)}
\end{overpic}
\hspace{0.1in}
\begin{overpic}[height=1.2in]{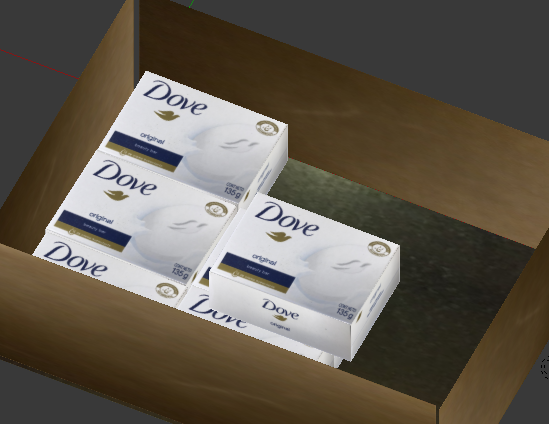}
\put(450,-70){(c)}
\end{overpic}
\begin{overpic}[height=1.2in]{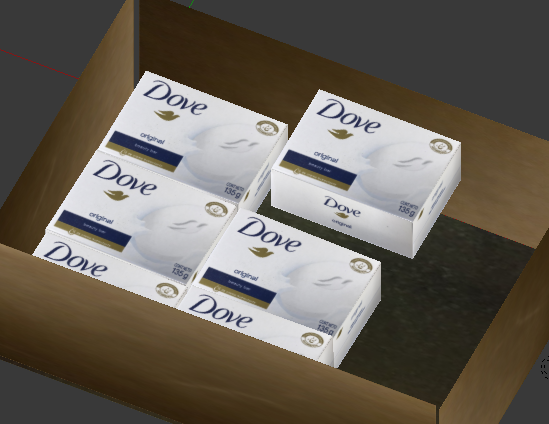}
\put(450,-70){(d)}
\end{overpic}
\begin{overpic}[height=1.2in]{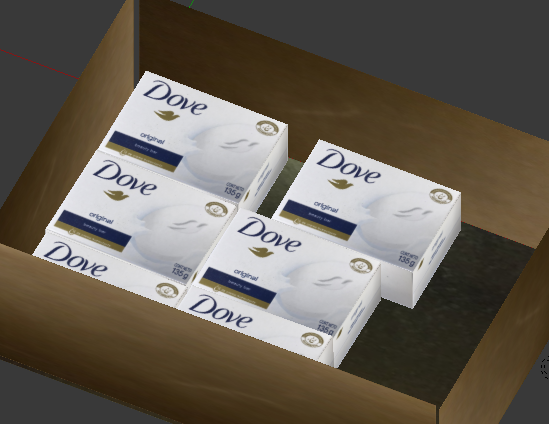}
\put(450,-70){(e)}
\end{overpic}
\caption{Adaptive pushing: (a) The black border is the bin and the gray rectangle is a previously placed object. The green rectangle represents the target pose for the current object. The light green boundary represents an $\varepsilon$-expanded model of the target object that intersects the point cloud at the purple points. These points result in the black vector that pushes the object away from them. (b) A screen shot of a scene's point cloud. The white points are collision points with previously placed objects. The red volume shows the computed pre-push pose for the new object. (c) The rest of the images show adaptive pushing for placing an object at the top-right corner, where there is no free space given the bin. (d) The target object is pushed against the boundary and then pushed down to the level of the objects. (e) The target object is moved towards its target pose at the bottom level.}
\label{fig:adaptive-pushing}
\vspace{-0.25in}
\end{figure*}

The {\sc GetOffset} primitive is described in the second image of Fig~\ref{fig:toppling_steps}. Given the pose $\pose_{\mathrm plane}$ of the object, the height of the top face \textit{z}, and the dimension of the edge(s) that are not adjacent to the desired face \textit{s}, define $\theta$ as $\tan^{-1} \frac{s}{z} $. The toppling operation needs to change the orientation of the object such that its center of gravity moves out of the support point along the contact edge denoted by the white circle. This object rotation can be achieved by displacing the pick's point of contact (green circle) to the {toppled configuration} (red circle). This offset is computed as:

\vspace{-0.2in}
$$
	\Delta x = \frac{s}{2} + z \sin \theta - \frac{s}{2} \cos \theta + \epsilon;\ 
	\Delta z = z \cos \theta + \frac{s}{2} \sin \theta - z.
$$
\vspace{-0.15in}

Here, $\Delta x$ and $\Delta z$ describe the motion vector for the end-effector shown as a red arrow in the third image of Fig~\ref{fig:toppling_steps}.

\subsection{Point Cloud Driven Adaptive Pushing}

Directly placing the $i^{th}$ object at the target pose $\hat{\pose}^i$ is prone to placement failures. Previous placement errors can cause surrounding objects to block $\hat{\pose}^i$. Executing a direct placement may result in damaging the objects. This means that the object can only be lowered safely to an offset pose that avoids surrounding objects. Once lowered to this offset, ''pre-push" pose, a key observation is that \textit{pushing in the direction of the target pose} can a) enable the current object to reach its target pose with low error, and b) correct local errors through compliant interactions with other objects. These observations motivate the proposed \textit{adaptive pushing primitive}. 

In order to compute a collision-free "pre-push" pose $\pose_{pre}$ the method operates over a point cloud snapshot $X$ taken of the target bin with all the previously placed objects. The cuboidal model of the object $B_{box}$ is evaluated \textit{in simulation} at the target pose $\hat{\pose}^i$. Due to errors in prior placements, parts of $X$ intersect with $B_{box}$. The set of intersecting points can be used to compute a planar offset direction to reduce collisions. Doing this repeatedly by offsetting the pose with small planar displacements will succeed once the offset pose has no intersection with $X$. Once computed, the push direction $\vec{d}$ takes the object from the safe $\pose_{pre}$ to target $\hat{\pose}^i$. 

There is another source of error arising from sensing and calibration. This means that $\pose_{pre}$ is collision-free only as long as the detected object pose and current execution is perfect. Real-world noise means that attempting to lower the object at $\pose_{pre}$ will encounter errors that cause the ground-truth object pose to differ from $\pose_{pre}$. Expanding the planar dimensions of the cuboidal object by $\epsilon$ to compute $B_{box}$ means that as long as the noisy ground-truth is contained within the expanded $B_{box}$ the real-world execution is \textit{still guaranteed to be collision free}. Similarly, a small raise in height $h$ is also incorporated to account for vertical noise, so that the pushing action is clear of the floor of the container. Fig~\ref{fig:adaptive-pushing} (a,b) is an illustration of the offset computation.

Algorithm~\ref{algo:PushPlace} outlines a single {\sc AdaptivePush} primitive. Line 3 generates the bounding box region with expanded object dimensions at a raised target pose. The loop between Lines 4-8 iteratively checks for neighboring \textit{collisions}, and displaces the pre-push pose by a step $\vec{u}$. Line 8 updates the bounding box. The primitive returns pre-push pose $\pose_{pre}$ and push vector $\vec{d}$. Certain implementation details are important to consider. The magnitude of $\vec{u}$ is kept sufficiently small ($5mm$) and there is a local minimum check if no progress is made, which can shrink $\epsilon$ if needed. 

It should be noted that a robust {\sc AdaptivePush} implies there exists a clearance of $| \vec{d} |$ from the target pose away from the neighboring objects. If the dimensions of the container are exactly (or close to) the dimensions of the packing arrangement, such a clearance will not exist. An alternative outlined in Algorithm~\ref{algo:TightPushPlace} leverages the compliance of the container walls to solve the low-clearance case.

\noindent
\begin{minipage}{0.48\textwidth}
\begingroup
\removelatexerror
\begin{algorithm}[H]
\small
\DontPrintSemicolon
\KwIn{Pointcloud $X$,  target pose $\hat{\pose}^i$, expansion $\varepsilon$, raise height $h$, object $o$}
\KwOut{Push vector $\vec{d}$, pre-push pose $\pose_{pre}$  }
$\pose_{pre} \gets $ {\sc Raise}$(\hat{\pose}^i, h)$ \;
$S\gets ${\sc Expand(Dimensions}($o$) $, \varepsilon$)\;
 $B_{box} \gets$ {\sc GenerateBoundingBox}($\pose_{pre}, S$) \;
\While{$B_{box} \cap X \neq \emptyset$ \textbf{or} $MaxIters$ }{
$X_{collision} \gets B_{box} \cap X$ \;
$\vec{u} \gets$ {\sc GeneratePushVector}($X_{collision}, \pose_{pre}$)\;
$\pose_{pre} \gets \pose_{pre} + \vec{u}$\;
$B_{box} \gets$ {\sc GenerateBoundingBox}($\pose_{pre}, S$) \;
}
$\vec{d} \gets ${\sc Raise}$(\hat{\pose}^i, h) - \pose_{pre}$\;
\Return{($\vec{d}, \pose_{pre}$)}\;
\caption{{\sc AdaptivePush}}
\label{algo:PushPlace}
\end{algorithm}
\endgroup
\begingroup
\removelatexerror
\begin{algorithm}[H]
\small
\DontPrintSemicolon
\KwIn{Pointcloud $X$,  target pose $\hat{\pose}^i$, expansion $\varepsilon$, raise height $h$, container height $H$, object $o$}
\KwOut{Tight Push-place primitive $\Pi$ }
$\pose^{high} \gets ${\sc Raise}$( \hat{\pose}^i, H$)\;
$\Pi \gets$ {\sc AdaptivePush}$(X, \pose^{high}, \varepsilon, 0, o)$  \;
$\Pi \gets \Pi\ \cup\ ${\sc AdaptivePush}$($X$, \hat{\pose}^i, \varepsilon, h, o)$\;
\Return{$\Pi$}\; 
\caption{{\sc TightPushPlace}}
\label{algo:TightPushPlace}
\end{algorithm}
\endgroup
\end{minipage}

\noindent\textbf{Adaptive Pushing in Low-clearance Packing: }
Algorithm~\ref{algo:TightPushPlace} outlines how repeated calls to Algorithm~\ref{algo:PushPlace} can achieve packing in tight containers. $\hat{\pose}^i$ in such a case might have container walls surrounding it.
The single push-primitive is first called on Line 2 for the object at the target pose raised by the container height $H$, which is clear of the surrounding objects, but still alongside the container wall. For poses along the edge of the container the first step pushes against the wall(s). For poses near the boundary, as in Fig.~\ref{fig:adaptive-pushing} (c,d,e), the first displacement (Line 2) pushes \textit{into the walls of the container}. If the container is compliant enough this push can take the object away from local errors for safely lowering to the $\pose_{pre}$ (computed on Line 3), and then the planar push towards the final pose $\hat{\pose}^i$.

\begin{figure*}[ht]
\centering
\begin{tabular}{@{}c@{}c@{}}
& No corrective actions\\
\multirow{3}{*}{\includegraphics[width=0.24\textwidth]{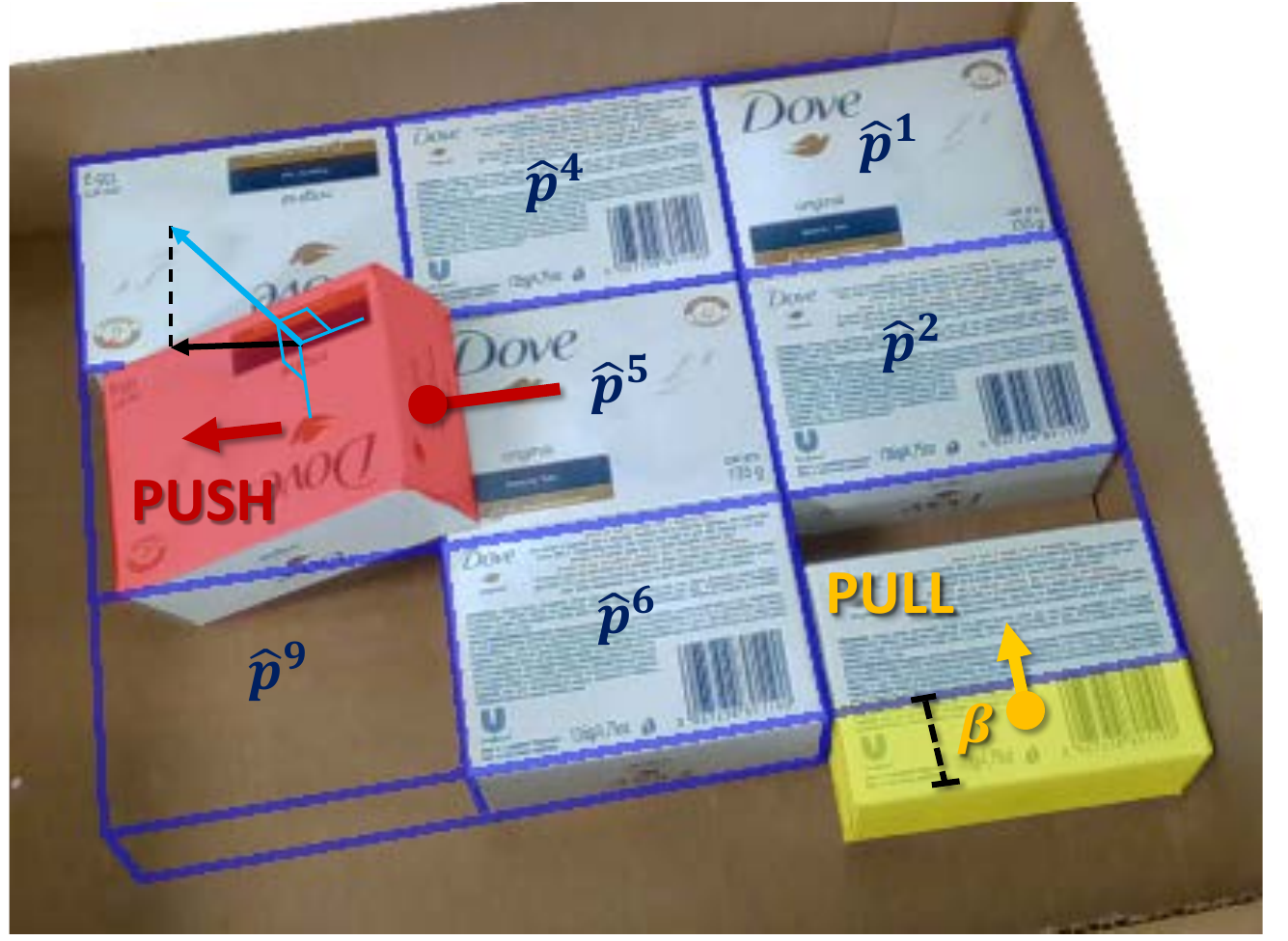}}
& \includegraphics[width=0.75\textwidth,valign=t]{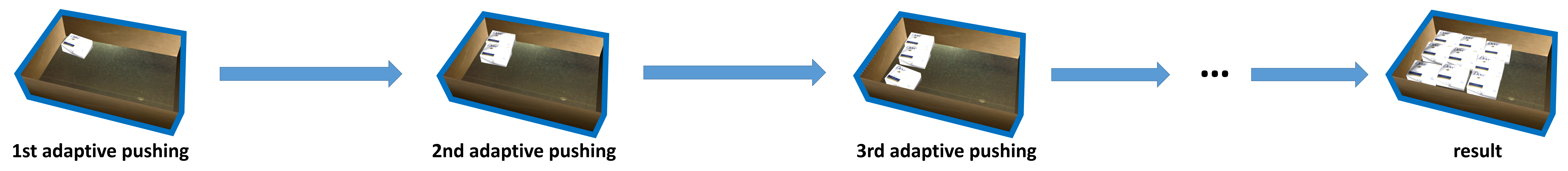}
\\\cline{2-2}
& Full pipeline\\
& \includegraphics[width=0.75\textwidth]{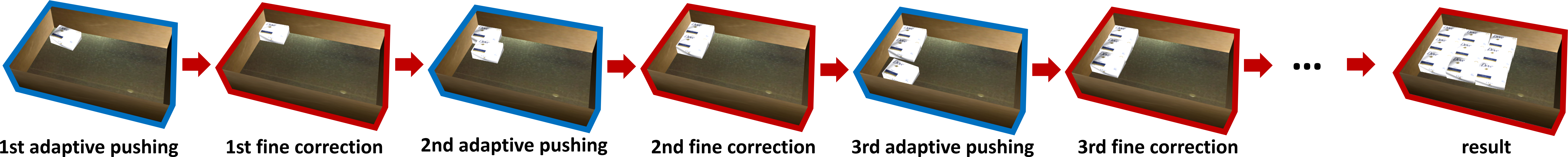}
\\
\end{tabular}
\caption{ (\textit{Left:}) The two kinds of errors that invoke fine corrections. (\textit{Right:}) Alternating adaptive push, and fine corrections for packing an object sequence. }
\label{fig:finer_fix}
\vspace{-0.2in}
\end{figure*} 
\subsection{Fine Correction using Push and Pull Primitives}
The final primitive deals with the remaining failure cases. Fine corrections are required because objects can be placed in incorrect poses due to unexpected collisions as well as calibration and pose estimation errors.  The proposed corrective manipulation procedure continuously monitors the scene and triggers corrective actions whenever necessary. 

\begin{minipage}{0.48\textwidth}
\begingroup
\removelatexerror
\begin{algorithm}[H]
\DontPrintSemicolon
\KwIn{
	$\atarget = \{ \hat{p}^1, \ldots, \hat{p}^n \}$, threshold $\tau$, corner point $x_{pivot}$, support surface normal $\vec{N}_s$
}
\Repeat{NoError \textbf{or} Timeout}
{
    $\{X^i\}_{i=0}^{n} \gets ${\sc TopFaceSegments}()\\
    {\scriptsize\tcp{Sort centers of $X^i$ based on increasing distance to $x_{pivot}$}}
    $\mathcal{X} \gets ${\sc SortByDistance}($\{X^i\}_{i=0}^{n}, x_{pivot}$)

    \For{$X^i \in \mathcal{X}$}
	{
		{\scriptsize\tcp{Check if $X^i$ is not horizontally placed}}
		\If{ $\exists x\in X^i,\ {normal(x)}{\boldsymbol{\cdot}}\vec{N}_s \leq 1 - \epsilon$ }
		{
		$x_c \gets center(X^i)$\\
		{\scriptsize\tcp{Push segment $X^i$ from a side to translate its center $x_c$  after projecting vector $normal(x_c)$ to the XY plane, and where $\alpha$ is a small constant}}
		$\Pi \gets ${\sc ProjectXY}$(normal(x_c), \vec{N}_s)  \times \alpha $\\
		{\sc Execute}$(\Pi)$\\
		\textbf{break}
		}
		{\scriptsize\tcp{Check if X[i] is horizontally offset}}
		$\beta \gets ${\sc Hausdorff}$( \hat{p}^{i}, X^i )$\\
		\If{$\beta > \tau$}
		{
		    {\scriptsize\tcp{Pull $X^i$ toward $\hat{p}^{i}$ with distance $\beta$}}
			$\Pi \gets ${\sc PullTowards}$(X^i, \hat{p}^{i}, \beta)$\\
			{\sc Execute}$(\Pi)$;
			break
		}
	}
}
\caption{\sc FineCorrection}
\label{algo:FinerFix}
\end{algorithm}
\endgroup
\end{minipage}

The proposed realignment process is illustrated in Algorithm~\ref{algo:FinerFix}. The process first removes the background, the box, the robot's arm, end-effector, and the side-faces of the objects from the observed point cloud. The remaining  point cloud corresponds to only top surfaces of objects, and is segmented into disjoint subsets $\{X^i\}_{i=0}^n$, where $X^i$ is the point cloud corresponding to object $i$. 
The observed point cloud is then compared against the desired alignment of the objects in their target poses $\{ \hat{p}^1, \ldots, \hat{p}^n \}$. 
The algorithm iterates through the objects in the order of their increasing distances to a corner point $x_{pivot}$. Misalignment is systematically detected by comparing the observed point cloud $X^i$ of each object to its desired one $\hat{p}^i$. 
As shown in Fig.~\ref{fig:finer_fix}, two types of misalignment errors can occur. 
The first type occurs when the top surface normal of an object, denoted by ${normal}(x_c)$,  is not perpendicular to the support surface (line 5). This error is corrected by pushing the object along a direction and for a distance given as the projection of the  surface normal on the support surface. This process is repeated until the surface normal becomes perpendicular to the support surface. 

The second type of error happens when a peripheral object is not entirely within the desired footprint of the pile (line 11). The proposed procedure systematically detects pivot points that are outside the desired footprint of the pile and pulls the objects inside accordingly (line 12). The correction is repeated until the point cloud is  aligned with the desired goal poses, within a given threshold $\tau$, or a timeout occurs.

\section{Simulation Framework}
\label{section:simulation}
\begin{figure*}[t]
\centering
\includegraphics[width=\textwidth]{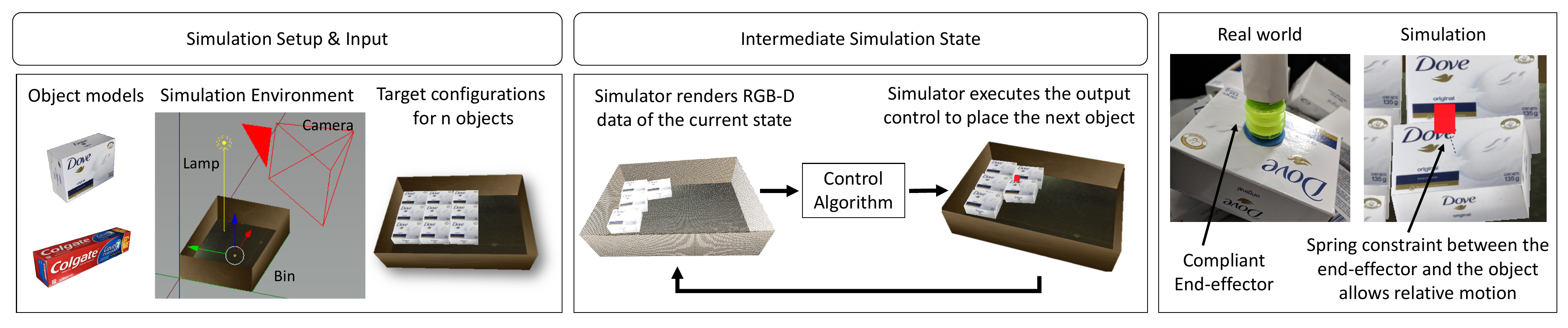}
\vspace{-.3in}
\caption{The simulation environment is set up with a bin, lights and camera parameters. Objects are placed sequentially to achieve a desired target configuration. Before each placement, the state of the bin is captured by rendering RGB-D images. The state is provided as input to the control algorithm that computes the motion for the target object. Finally, the controls are executed in the simulator to place the object in the bin. To accurately model the compliance of the real-world end-effector, a spring constraint is applied in simulation between the end-effector and the object. The constraint allows small relative motion and prevents unrealistic collision and jumping of objects that are often observed in rigid-body simulation.}
\label{fig:simulator}
\vspace{-.2in}
\end{figure*}

A simulation framework is developed to model the physical aspects of real-world tight packing. Given the focus of developing and evaluating sensor-based control algorithms, such as {\it adaptive pushing} and {\it fine-correction}, the framework simulates an RGB-D sensor to capture the state of the process. Fig.\ref{fig:simulator} shows the simulation setup in Blender\footnote{\url{http://www.Blender.org}}. 

The software framework loads a model of the {\it Bin} into the simulation environment. A {\it point lighting source} and a {\it camera} are initialized based on the parameters specified in a configuration file. A target packing sequence is specified for objects with known 3D models as $\{ (o^i, \hat{\pose}^i) | i=1:n\}$, where $o^i$ identifies the object and $\hat{\pose}^i$ denotes the desired target pose of the object. An example of the desired configuration is shown in Fig.\ref{fig:simulator} (left). 

An intermediate simulation state corresponds to the partial placement of objects in the bin. The achieved placement might be inaccurate due to the noise (artificially added in the simulation framework) from execution or sensing. The resulting state is captured via an RGB-D rendering of the scene. The simulation can either be set to render images via a fast rasterization-based rendering process or to generate photo-realistic images via physically based rendering. In addition, the framework generates depth images, instance segmentation masks and consequently the 3d point cloud data. This state information is provided as  input to {\it control algorithms}, such as adaptive pushing or fine correction. The control algorithms generate motions of the end-effector for the placement of the target object which are then physically simulated via the Bullet Physics Engine\footnote{\url{http://www.bulletphysics.org}}. 

The simulator reasons about the physical interactions between the end-effector and the object being manipulated as well as the interactions between different objects. The simulator is developed to mimic the compliance of the real-world setup as shown in Fig.\ref{fig:simulator} (right). This is achieved by introducing a {\it spring constraint} between the end-effector and the object being manipulated. The spring constraint allows the link between the end-effector and the object to be stretched during the packing operation, which is similar to the behavior achieved by the soft suction gripper. In the absence of this constraint, i.e. in a strictly rigid simulation, the objects would pop out to resolve the collision.

The simulation framework is publicly released with this work and can be used for two purposes. The first is for developing and evaluating sensor-based control algorithms as in the case of the current work. This allows extensive evaluation and ablation studies of algorithms and environments with varying parameters,  as well as a variety of objects and configurations without the need for an expensive robotic setup. The second use-case is in the context of a learning-based control algorithms, such as reinforcement learning, where this framework can be used to generate a large-scale training dataset. The framework gives access to photo-realistic image rendering as well as 3D point cloud data. It also generates corresponding instance masks and pose labels. This state information can be associated with the corresponding action sequences. Furthermore, the framework allows domain randomization over different parameters of the environment, which can help transfer learning and deployment in real-world setups.

\section{Evaluation}
\label{section:evaluation}
To evaluate the performance of the proposed pipeline, extensive experiments were executed both in simulation and on a real robotic platform, which encompasses the motivating complications that arise in real-world setups.

\begin{figure*}[ht]
\centering
\begin{minipage}{0.7\textwidth}
\centering
  \includegraphics[width=\textwidth, height=1.7in]{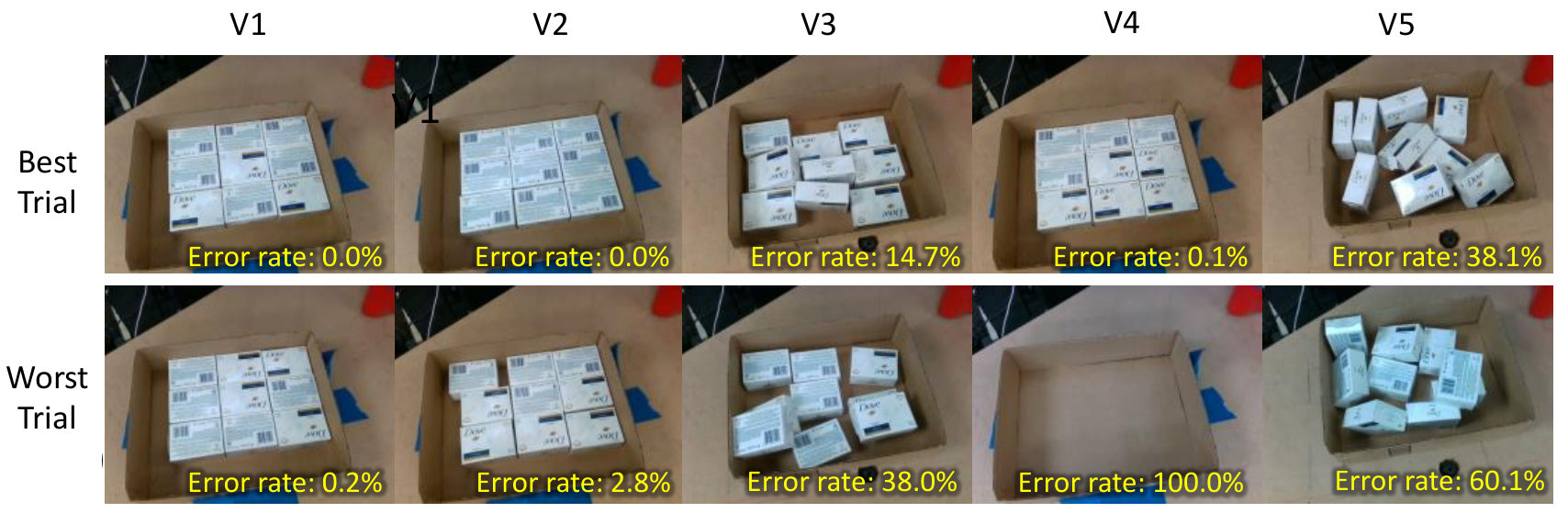}
\end{minipage}
\begin{minipage}{0.29\textwidth}
\vspace{0.09in}
\centering
  \includegraphics[height=0.7in, trim={0.4in 0 0.5in 0},clip]{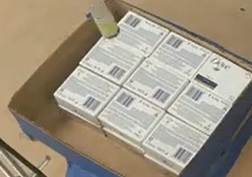}
  \includegraphics[height=0.7in, trim={0 0 0.5in 0},clip]{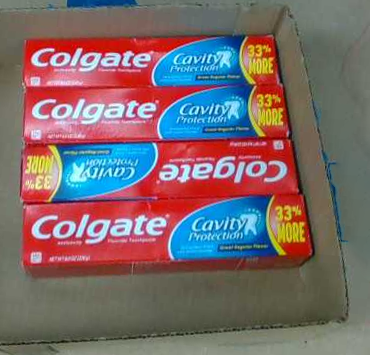}
  \includegraphics[height=0.7in, trim={1.3in 0 1in 0},clip]{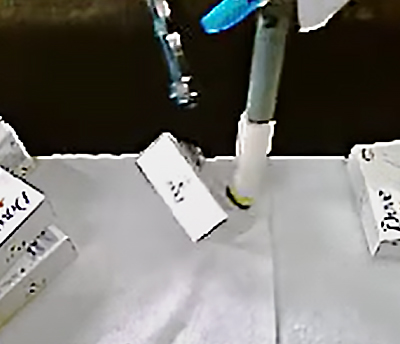}\\
  \vspace{0.05in}
  \includegraphics[height=0.7in, trim={0in 0 0 0},clip]{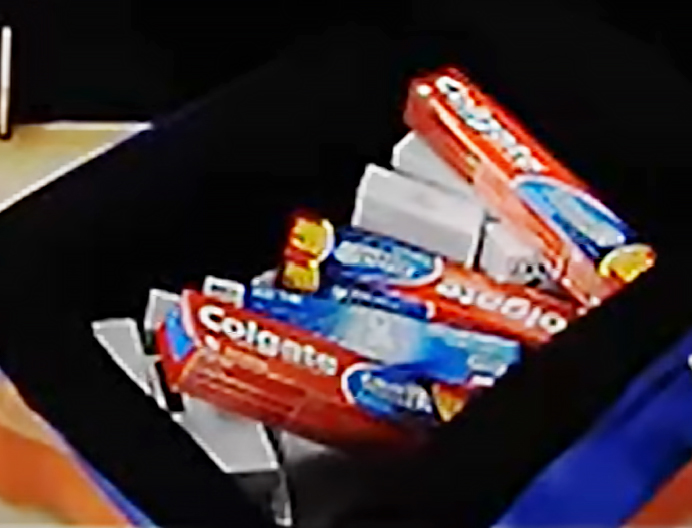}
  \includegraphics[height=0.7in, trim={0 0 0in 0},clip]{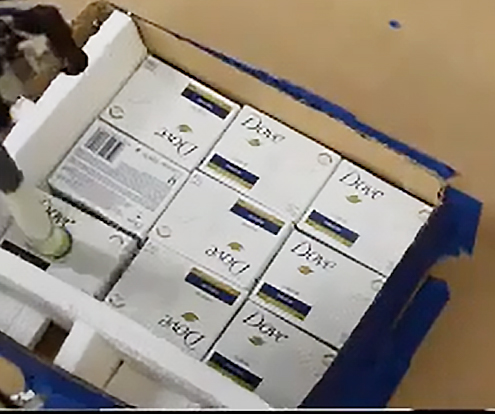}
\end{minipage}
\vspace{-0.15in}
\caption{The final set of object poses in the target bin at the end of every experiment. Different columns represent different versions. The \textit{Left top row} is the best case, and the \textit{Left bottom row} is the worst case. (\textit{Right:}) Demonstrative real-world trials performed with the same pipeline for (\textit{from top left}) multi-layer packing, different objects, narrow-face toppling, heterogeneous piles, and no-clearance packing.}
\vspace{-0.15in}
\label{fig:final_bin_configs}
\end{figure*}

\subsection{Hardware and Software Setup}
\label{section:components}
First, this section describes the hardware setup, the workspace design and software choices, which influence the baseline pipeline and the proposed manipulation primitives:

\paragraph{Hardware Setup:} The robot is a $\mathtt{Kuka\ IIWA14}$ $7$-DoF arm (Fig.~\ref{fig:new_setup} middle). A customized end-effector solution extrudes a cylindrical end-effector that ends with a compliant suction cup. Two \textit{RealSense} \commentadd{SR300} cameras are mounted on a frame and pointed to the $\binit$ and $\btarget$ containers from the other side relative to the robot as Fig.~\ref{fig:new_setup} (right) shows. The cameras' frame is statically attached to the robot's base such that calibration errors are minimized in estimating the camera poses in the robot's coordinate system.

\paragraph{Workspace Design:} 
Fig.~\ref{fig:new_setup} (right) shows the setup designed for the target task. The annular blue region represents the subset of the reachable workspace that allowed for top-down picks with the robot's end-effector. This region is computed by extensively calling an inverse kinematics (IK) solver for top-down picks with the end-effector. The IK solutions indicate that the radial region between $40cm$ and $70cm$ from the robot center maximizes reachability and IK solution success given the setup.  The bins (red rectangles) are placed so that they lie inside the optimal reachable region.

\paragraph{Software Dependencies}
{\tt MoveIt!}~\cite{chitta2012moveit} is used for motion planning. Most of the motions are performed using \textit{Cartesian Control}, which guides the arm using end-effector waypoints. Ensuring the motions occur in reachable parts of the space increases the success of \textit{Cartesian Control}, simplifies motion planning and speeds-up execution. To decrease planning time, motion between the bins is precomputed using $\mathtt{RRT^*}$~\cite{Karaman2011Sampling-based-} and simply replayed at appropriate times.

\subsection{Experimental Design}
\label{section:experimental_design}

The experiments are designed to showcase the hardness of tight packing as well as the benefits of adding robust environment-aware manipulation primitives that aid in increasing success rate and accuracy. To evaluate the performance of the primitives over different noise levels and to study the sensitivity of the primitives with respect to the underlying parameters, extensive experiments were executed on a physics-based simulator. The output of {\it real-world experiments} and {\it simulation experiments} are evaluated over three distinct metrics:

\textbf{Volumetric Error:} This measures the percentage of unoccupied volume within the ideal target placement volume. This is measured by computing an occupancy grid based on the sensing data acquired after the objects have been moved to the target bin. Given the partial observation, an approximation of the volume is computed by projecting the sensed points down to the bottom surface of the bin and marking all voxels along the projection as occupied. 

\textbf{ADI Error \cite{hinterstoisser2012model} : } This is often used in the pose estimation literature. Given two poses, the ADI metric measures the average distance between corresponding points on the object model placed at the two poses. 
 
\textbf{Pose Recall:} A pose is assigned to each of the object placements by aligning a 3D model of the object to the observed point cloud data. This pose is then compared to the closest target pose to compute an ADI error. A pose is considered correct if the ADI error is less than a distance threshold. Pose recall curves indicate the number of successful object placements that are within different threshold margins.

\begin{figure*}[h]
\centering
\includegraphics[width=0.95\textwidth]{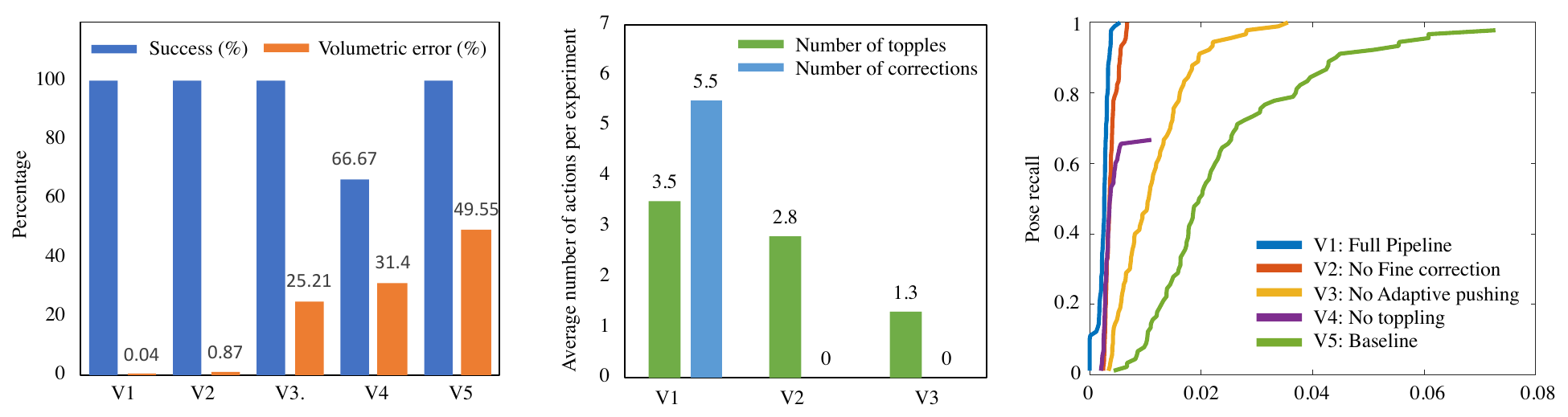}
\vspace{-0.15in}
\caption{(\textit{Left}): The blue bars represent the fraction of successful object transfers. The orange bars represent the percentage of unoccupied volume within the ideal target placement volume. (\textit{Middle:}) The blue bar represents the average number of correction actions happened per experiment. The green bars represent the average number of toppling actions occurred per experiment. (\textit{Right:}) The pose recall percentage over different thresholds is shown for the pipeline versions.}
\vspace{-0.15in}
\label{fig:real_data}
\end{figure*}

\subsection{Real-world Packing Trials}
For consistency, an identical version of the problem is tested, with ``dove soap bars'' that are randomly thrown into the source bin $\binit$, which is placed on one side of the robot's reachable workspace (Fig \ref{fig:new_setup}). Only top-down grasps are allowed within a given alignment threshold. The start arrangement $\ainit$ of objects is intended to reflect a random pile, with $10$ repetitions of each experimental condition. The target bin $\btarget$ contains a $3\times 3$ grid arrangement of $9$ objects, on the same plane, with the stable face of the object targeted for placement. The complete pipeline uses a) corrective actions for fine adjustments, b) push-to-place actions for robust placement, c) toppling actions for increasing successes, and d) pose estimation for adjusting the object. The improvements introduced by these strategies are evaluated through the following comparison points, within the context of the proposed pipeline: 

\textbf{V1 - \commentadd{Full pipeline}}: The complete pipeline with all the primitives achieves the highest accuracy and success rate.

\textbf{V2 - No corrective actions}: The experiment corresponds to \textbf{V1} without the fine correction module of Fig.~\ref{fig:finer_fix}.

\textbf{V3 - No push-to-place actions}: This version is \textbf{V2} without the use of the robust placement module (Fig. \ref{fig:adaptive-pushing}) that performs push actions to achieve robust placement.

\textbf{V4 - No toppling actions}: These experiments used \textbf{V2} without considering toppling actions to deal with objects not exposing a valid top surface that allows the target placement. 

\textbf{V5 - (Baseline) No push-to-place, toppling, pose-estimation}: The naive baseline that solely uses a pose-unaware grasping module that reports locally graspable points and drops the grasped object at an end-effector pose raised from the center of the desired object position, with no adjustment in orientation. 

The metrics evaluated include the fraction of successful object transfers that succeed in moving objects to the target bin. The accuracy is captured in the threshold mentioned in Eq.~\eqref{eq:satisfaction} that is expressed in terms of a percentage of unoccupied volume within the ideal target placement volume. This was measured with a voxel discretization sufficient to elucidate the difference between the methods. The average data recorded is reported in Fig.~\ref{fig:final_bin_configs} (\textit{left}). This error measure is proportional to the accuracy. The key observations from these experiments, regarding each variant, are detailed as follows.

\begin{figure*}[t]
    \centering
    \includegraphics[width=0.9\textwidth]{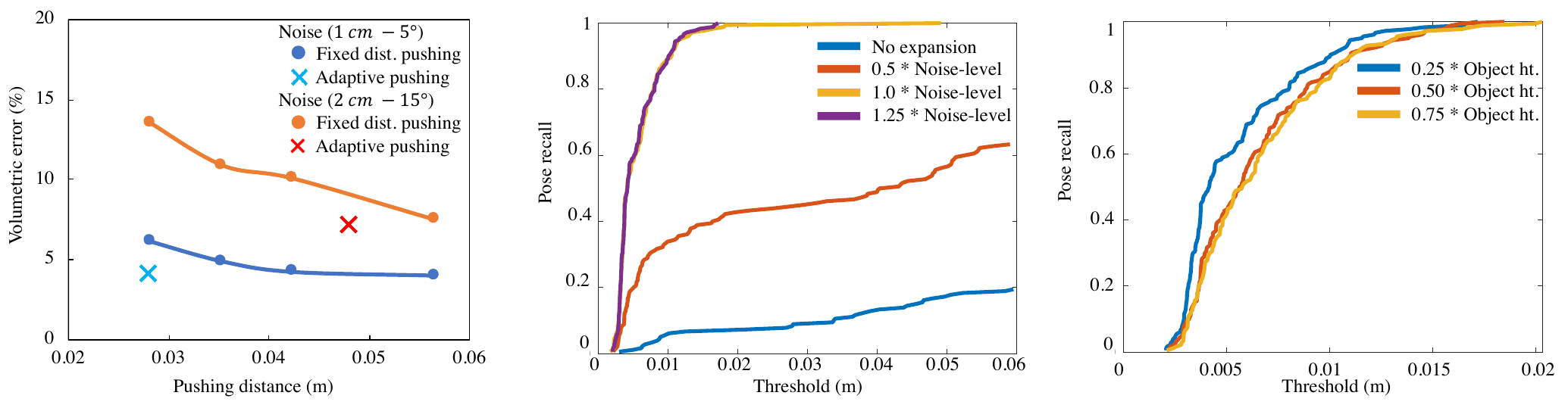}
    \vspace{-0.15in}
    \caption{(\textit{left}) Volumetric error (ADI) of proposed adaptive pushing primitive compared with baseline method under two noise levels. Pushing distance is a parameter for the baseline method. 
    Pose recall against the parameter for adaptive pushing. (\textit{middle}) pose recall against the expansion size of object model, when the expansion size is above the noise level, the performance is similar; as the expansion size decreases below noise level, the performance becomes worse. From the 0 noise level to 1.25 times noise level, the successful rates are $7.2\%, 42.2\%, 99.4\%, 100\%$,  \textit{(right)} pose recall against height of pre-push pose, it shows that the performance drops rapidly after the height of pre-push pose increases over 1/4 of object height.}
    \vspace{-0.15in}
    \label{fig:adaptive-pushing-compare}
\end{figure*}

The low error for \textbf{V1} corroborates the final bin placement evidence. On average, seven  corrective actions per experiment were invoked to achieve the high degree of accuracy.

The accuracy improvement obtained from corrective actions is evaluated in \textbf{V2}. While this version succeeded in dropping all the objects close to the correct target poses, application use-cases where a higher degree of accuracy is desired motivate the use of corrective actions. The integration of the corrective actions was done with higher error threshold during intermediate steps, and a much finer one for the final adjustment. Errors can typically arise from execution failure and pose misalignments. The less accurate these underlying processes are, the more important corrective actions become. 

\textbf{V3} only performs adjustments using pose estimation, and toppling. While, this is sufficient to successfully transfer all the objects, any difference of accuracy to \textbf{V2} would be introduced by the lack of push-to-place actions. Here, there is complete reliance on the exactness of the execution and pose adjustments. Due to the proximity of adjacent object surfaces in the target grid arrangement, even minor errors get aggravated. However, due to the ability to reason about toppling, all the objects can be transferred to the target bin, even with this low accuracy. This is demonstrated in the occurrence of the failure to transfer all the objects. 

In \textbf{V4}, any object that does not expose a permitted picking surface that makes the prehensile placement possible, is not picked. Any instance of the source bin that ends with no such objects results in no valid picking actions that can make the approach proceed, and a failure is thus declared. The current behavior of \textbf{V4} drops the object if it is mistakenly grasped from the wrong surface. This can itself be used as a naive toppling primitive. It is important to note that there might be other alternative strategies that can deal with this failure, but the intent of this comparison is to demonstrate the importance of having a deliberate toppling strategy in the pipeline, that can change the object's orientation in the context of random starting arrangements of the object.  On average, the toppling primitive was required four times per experiment in \textbf{V1},  \textbf{V2}, and \textbf{V3}. This highlights the necessity of this reasoning. Deliberate toppling however requires at least one additional pick action. The number of pick attempts per successful object transfer was $2.56$ for \textbf{V4}, whereas, in \textbf{V2} the same was $1.98$. This indicates that toppling is indeed necessary both in terms of success and efficiency of actions.

\begin{figure*}[t]
    \centering
    \includegraphics[width=0.9\textwidth]{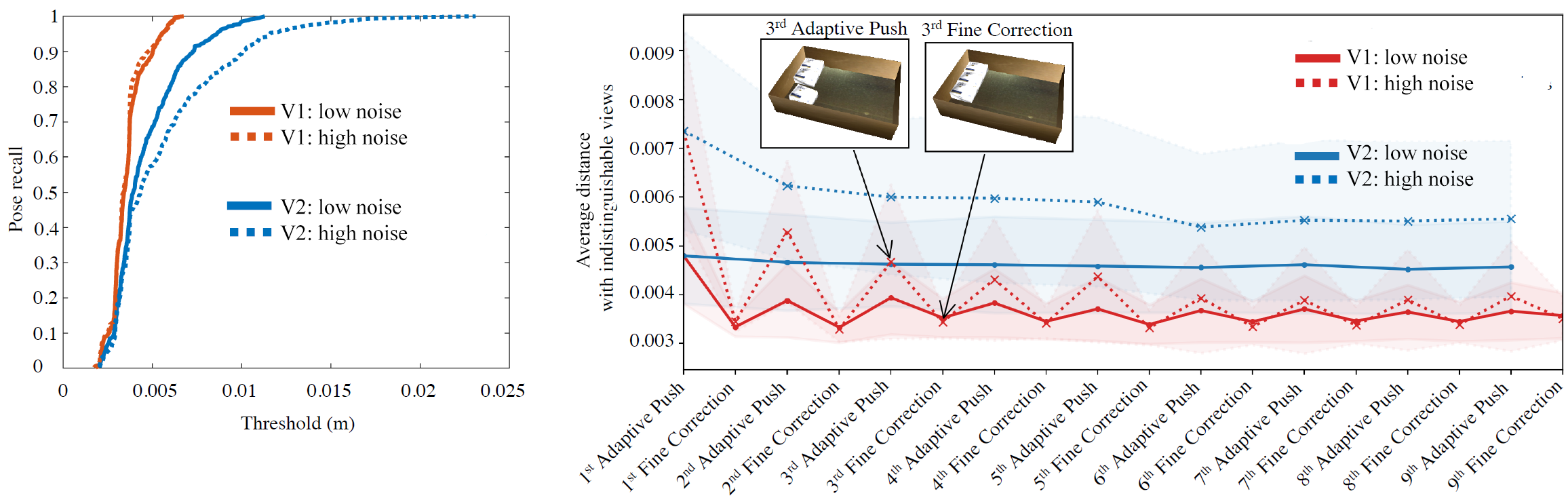}
    \vspace{-0.15in}
    \caption{Pose recall (left) and average distance error (ADI) against the number of placed objects (right) with and without the fine correction module under different noise levels.}
    \vspace{-0.15in}
    \label{fig:res_recall}
\end{figure*}

Expectedly, \textbf{V5} has the lowest accuracy. However, since there is no reasoning about the pick surface, every object was transferred to the space of the bin. This has no guarantee to work if the object is larger. This drives the motivation for using a set of robust primitives for the packing problem.

Overall, the time for the experiments shows a trend of increasing with the increasing complexity of the pipeline. The trade-off of accuracy versus time persists. On average, \textbf{V1} ran for $945$s per bin while \textbf{V5} ran for $323$s.

\subsection{Real-world Demonstrations}

Various demonstrative packing tasks, highlighted in Fig~\ref{fig:final_bin_configs} (\textit{right}) were executed to showcase the versatility of the proposed pipeline.

\textbf{Multi-layer packing}: Objects can also be packed in multiple layers. Running the method beyond the first plane of the grid shifts all the operations to the higher plane. This is demonstrated in a standalone run of the attached video.

\textbf{Different objects}: To validate the method's applicability to cuboid objects of different dimensions, \textbf{V1} was performed for toothpastes. Over $5$ experiments,  with $4$ objects, every run succeeded in placing the objects inside the bin.

\textbf{Narrow-face reorientation}: The packing orientation is tested with the narrow (unstable) face of the cuboid being the contact surface. Here toppling has to reorient the object from its more stable wider face to the desired narrow face. 

\textbf{Heterogenous pile}: The source bin was filled up with two different kinds of objects and the pipeline was executed for each object sequentially, i.e., pack the \textit{toothpaste} first, and then the \textit{soap}, into two separate containers. The method is correctly capable of retrieving, and packing the desired objects from such a heterogeneous pile.

\textbf{No-clearance Packing}: The target bin is resized to be of the exact dimensions as the cross section of the target arrangement. This leaves no clearance for placement along the edges and corners of the $3\times 3$ grid. The compliance of the end-effector, and the container's walls are leveraged by executing the meta-primitive of Algo~\ref{algo:TightPushPlace}.

\subsection{Ablation Studies in Simulation}
The simulator described in Sec~\ref{section:simulation} is used to perform ablation studies for both adaptive pushing, and fine correction. 

\textbf{Noise Level:} Real world trials introduce errors due to pose estimation, execution, calibration, and non-prehensile interactions. All of these cause a disparity between the objects' ground truth poses versus where the pipeline models it to be. The objective of adaptive pushing and fine correction is to account for this difference. In order to approximate the noise introduced by all of these sources of error, two noise levels are evaluated with the adjustment primitives: the lower noise level uniformly samples up to a $1cm$ planar translation offset and up to a $5^\circ$ deviation to the object's orientation given the ground truth. The higher noise level applies up to a  $2cm$ translation, and $15^\circ$ orientation. This means that the simulator changes the \textit{ground-truth} values given the noise level for perception purposes, while executing the manipulation operations over the unperturbed poses.

\textbf{Adaptive Pushing:} Simulated experiments examine the effectiveness of the proposed adaptive pushing primitive on reducing the packing error. The comparison is between the proposed method and a simple baseline method. For each new object to place, given the target pose, the baseline method calculates a pre-push pose using a \textit{constant} displacement. During the simulation, noise (sampled from either of the above two levels) is applied to the pre-push pose. In the experiment, the average volumetric error, and displacement computed from the proposed method is compared against the baseline method with different fixed displacements as parameters. The result is shown in Fig \ref{fig:adaptive-pushing-compare} (\textit{left}). The results show that as the pushing distance increases for the baseline method, its performance improves, and is better than adaptive pushing beyond $5cm$. Nevertheless, this means such a large clearance would be required between the dimensions of the container and the packing arrangement. Therefore the adaptive pushing, which achieves better performance for a lower average pushing displacement is more amenable to tighter packing scenarios.

To further evaluate the adaptive pushing primitive, the influence of the expansion parameter $\varepsilon$ from Algorithm~\ref{algo:PushPlace} is examined. The evaluation studies the effect of this parameter on the performance of the primitive. The result is shown in Fig. \ref{fig:adaptive-pushing-compare} (\textit{middle}). For the expansion size, different expansions relative to the maximum translation noise level and its effect on pose recall were studied.  It is observed that once the expansion size drops below the noise level, the pose recall curve drops drastically. The lower expansion curves achieve low success due to situations where the pre-push pose is in collision with neighboring objects, triggering a failure of that run (remainder from 9 objects). From the no expansion to 1.25 times the noise level, the success rate increases from $7.2\% $ to $100\%$. This motivates using a large enough expansion $\varepsilon$ that is larger than the estimate for the noise that errors in the system can introduce.

For the height of the pre-push pose ($h$), different values relative to the object's height were checked for how the pose recall changes. The result is shown in Fig. \ref{fig:adaptive-pushing-compare} (\textit{right}). The primitive is less sensitive to the value of $h$. Performance is best for a fraction (0.25) of object height.

\textbf{Fine Corrections: } The fine correction module was evaluated on the same simulation setup and noise levels.  The following alternatives were compared: \textbf{V1}, which includes adaptive pushing and the fine correction modules, and \textbf{V2}, which only performs adaptive pushing. Fine corrections are repeated at most 4 times per object until the point cloud is aligned with the desired goal poses within a threshold of $5mm$.
An expansion of $1.25$ times noise level was used since it achieved $100\%$ pose recall.
In this section, the focus is on how the fine correction module can improve the \textit{average distance error}. Fig~\ref{fig:res_recall} (left) shows the pose recall rate under different noise levels. In this experiment, an average of $1.03 (\pm0.80)$ corrective actions is executed per placement. Without the corrective module (\textbf{V2}), the pose recall declines substantially as the noise level increases. \textbf{V1}, however, preserves a high recall rate regardless of noise level. Fig~\ref{fig:res_recall} (right) shows the distance error (ADI) averaged over the number of placed objects. The error drops periodically after each execution of the fine correction module in \textbf{V1}, while the same error remains mostly unchanged when \textbf{V2} is used instead of \textbf{V1}. The fine correction module not only corrected the misplacement due to noise, but also prevented accumulative misalignment. The quality improvements increase with the increase in noise, thereby motivating fine corrections where larger amounts of noise is expected.

\section{Discussion}
\label{section:discussion}
The proposed pipeline indicates intriguing nuances of the packing problem. The use of a minimal, suction-based end-effector is a cost-effective, simple and relatively robust way to pick objects but does not easily allow for complex grasp reasoning, regrasping, or within-hand manipulation. The proposed pipeline achieves high accuracy by leveraging the compliance of the suction cup and the environment, while the object is attached. It uses robust reasoning to incrementally correct errors, instead of compounding them. While it can be argued that better baseline components can be developed to minimize uncertainty, the overall philosophy of robust, minimal, and compliant reasoning remains unchanged. 

The proposed system and primitives can also deal with cubic objects with different sizes and can be extended to non-cubic objects by adapting the object models. The key adaptation corresponds to identifying an appropriate packing arrangement $\atarget$ (potentially labeled in this case) in the target bin and the corresponding picking order from the initial bin. Future work will also explore speeding up performance and dealing with algorithmic challenges: the combinatorial reasoning over possible placements, physics-based reasoning to further improve pushing and toppling, as well as extending to more adaptive end-effectors. The platform can also be utilized as a training testbed for reinforcement learning to automatically discover robust primitives for solving packing tasks.

\bibliographystyle{IEEEtran}

\end{document}